\documentclass[sigconf]{acmart}

\usepackage{CJKutf8}
\usepackage{amsmath}
\usepackage{xspace}
\usepackage{graphicx}
\usepackage{algorithm}
\usepackage{algorithmic}
\usepackage{footnote}
\usepackage{makecell}
\usepackage{makecell}
\usepackage{booktabs}
\usepackage{multirow}
\usepackage{utfsym}
\usepackage{enumitem}
\usepackage{subfigure}
\usepackage{amsmath}

\AtBeginDocument{%
  }

\setcopyright{acmcopyright}

\copyrightyear{2023}
\acmYear{2023}
\setcopyright{acmlicensed}\acmConference[WWW '23]{Proceedings of the ACM Web Conference 2023}{May 1--5, 2023}{Austin, TX, USA}
\acmBooktitle{Proceedings of the ACM Web Conference 2023 (WWW '23), May 1--5, 2023, Austin, TX, USA}
\acmPrice{15.00}
\acmDOI{10.1145/3543507.3583279}
\acmISBN{978-1-4503-9416-1/23/04}

\begin{document}

\title{Meta-Learning Based Knowledge Extrapolation for Temporal Knowledge Graph}

%
\author{Zhongwu Chen}
\affiliation{%
  \institution{National University of Defense Technology}
  \city{Changsha}
  \country{China}
}
\email{chenzhongwu20@nudt.edu.cn}

\author{Chengjin Xu$^{\scriptscriptstyle *}$}

\affiliation{
  \institution{International Digital Economy Academy}
  \city{Shenzhen}
  \country{China}}
\email{xuchengjin@idea.edu.cn}

\author{Fenglong Su$^{\scriptscriptstyle *}$}
\affiliation{%
  \institution{National University of Defense Technology}
  \city{Changsha}
  \country{China}
}
\email{sufenglong18@nudt.edu.cn}

\author{Zhen Huang}
\authornote{Corresponding author}
\affiliation{ 
  \institution{National University of Defense Technology}
  \city{Changsha}
  \country{China}}
  \email{huangzhen@nudt.edu.cn}

\author{Yong Dou}
\affiliation{ 
  \institution{National University of Defense Technology}
  \city{Changsha}
  \country{China}}
  \email{douyong@nudt.edu.cn}

%
\renewcommand{\shortauthors}{Zhongwu Chen, Chengjin Xu, et al.}

\begin{abstract}
In the last few years, the solution to Knowledge Graph (KG) completion via learning embeddings of entities and relations has attracted a surge of interest. Temporal KGs(TKGs) extend traditional Knowledge Graphs (KGs) by associating static triples with timestamps forming quadruples. Different from KGs and TKGs in the transductive setting, constantly emerging entities and relations in incomplete TKGs create demand to predict missing facts with unseen components, which is the extrapolation setting. Traditional temporal knowledge graph embedding (TKGE) methods are limited in the extrapolation setting since they are trained within a fixed set of components. In this paper, we propose a Meta-Learning based Temporal Knowledge Graph Extrapolation (MTKGE) model, which is trained on link prediction tasks sampled from the existing TKGs and tested in the emerging TKGs with unseen entities and relations. Specifically, we meta-train a GNN framework that captures relative position patterns and temporal sequence patterns between relations. The learned embeddings of patterns can be transferred to embed unseen components. Experimental results on two different TKG extrapolation datasets show that MTKGE consistently outperforms both the existing state-of-the-art models for knowledge graph extrapolation and  specifically adapted KGE and TKGE baselines. \looseness=-1
\end{abstract}

\ccsdesc[500]{Computing methodologies~Knowledge representation
and reasoning; Temporal reasoning}

\keywords{Temporal Knowledge Graph Completion, Meta-Learning,  Knowledge Extrapolation}

\maketitle

\setlength{\parskip}{0.2cm plus4mm minus3mm}
\section{Introduction}
Temporal Knowledge Graph (TKG) is structured as a multi-relational directed graph where each edge
stands for an occurred fact.
TKG consists of a large number of facts in the form of quadruple (subject entity, relation, object entity, timestamp) , or (\emph{s}, \emph{r}, \emph{o}, \emph{t}) for short, where entities (as nodes) are connected via relations with timestamps (as edges). Many large-scale temporal knowledge graphs, including ICEWS~\citep{ICEWS}, YAGO3~\citep{YAGO3} and Wikidata~\citep{Wikidata}, have been successfully used in various knowledge-intensive downstream including question answering~\citep{saxena-etal-2021-question, Sun2020FaithfulEF, 10.1145/3184558.3191536, 10.1145/3269206.3269247}, 
recommendation systems~\citep{Wang2019MultiTaskFL,qertredd,asdfwef,Zhao2019CrossDomainRV}, and information retrieval~\citep{ zhang-etal-2019-long, 10.5555/3524938.3525667}.\looseness=-1
\par
Many TKGs are human-created or automatically constructed from semi-structured and unstructured text, suffering the problem of incompleteness, i.e. many missing links among entities. This weakens the expressive ability of TKGs and restricts the range of TKG-based applications. Many TKG embedding (TKGE) models like TComplEx~\citep{TComplEx}, TDistMult~\citep{TDistMult} and TeRo~\citep{TeRo} have been proposed to predict missing links by learning low-dimensional representations for a set of entities, relations and timestamps. These embedding-based methods obtain the rationality of the target quadruples by pair-wise distance and are widely used as score functions.\looseness=-1
\par

\par
\begin{figure}[!]
    \centering
    \includegraphics
    [width=\linewidth]
    {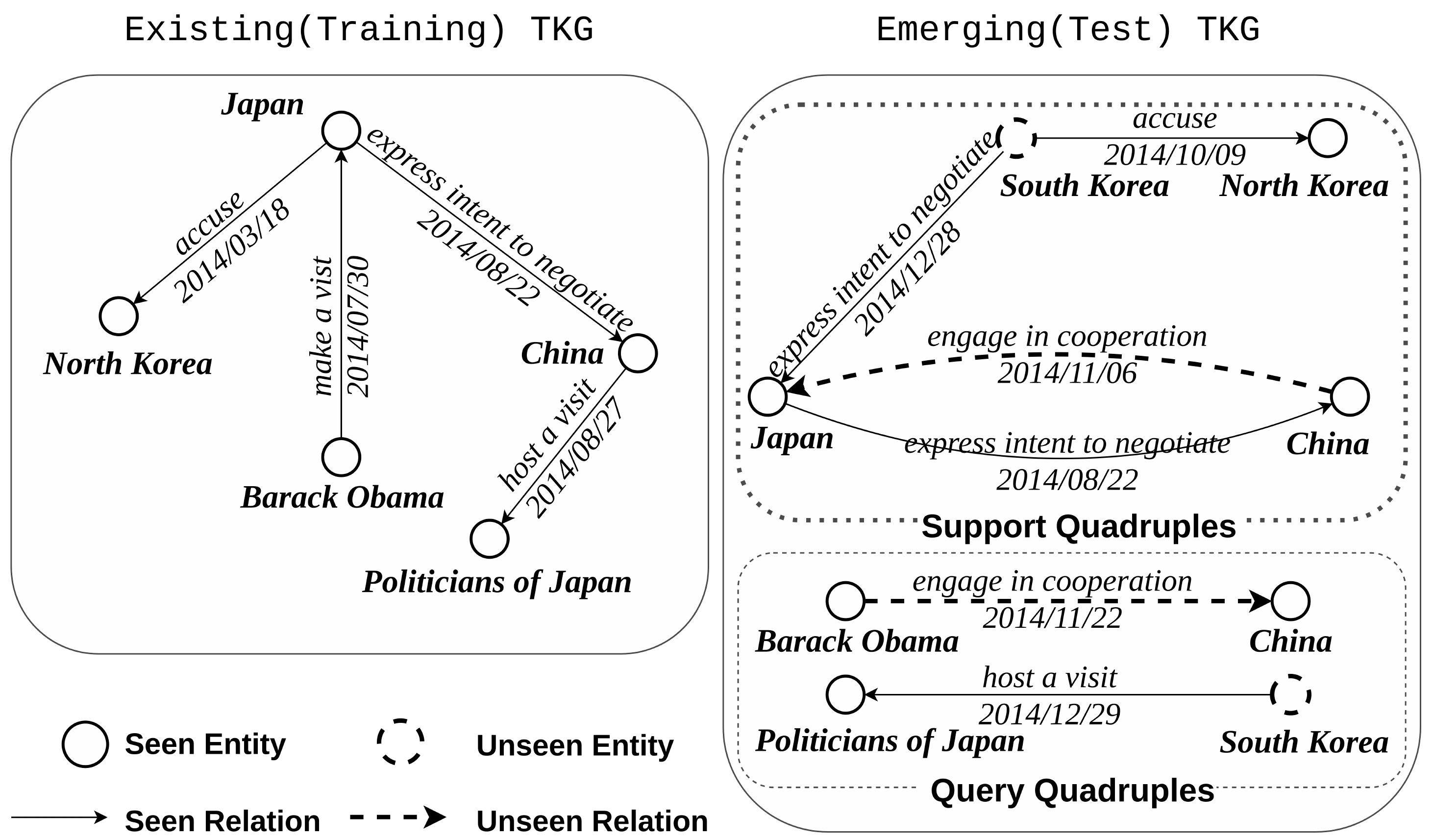}
    \caption{An example of TKG in the extrapolation setting. A new entity $ {South}$ $ {Korea}$ and a new relation $ {engage}$ $ {in}$ $ {cooperation}$ emerge in test TKG.}
    \vspace{-0.3cm}
    \label{fig1}
\end{figure}

However, TKGs are developing rapidly, where new entities and new relations arise along with time, and emerging TKGs are also incomplete.  Taking the case in Fig.~\ref{fig1} as an example, the emerging TKG contains an unseen entity $ {South}$ $ {Korea}$ and an unseen relation $ {engage}$ $ {in}$ $ {cooperation}$; traditional TKGE models trained on the existing TKG can not obtain the embeddings of the two components, so they can not answer the two missing links in Query Quadruples of the emerging TKG. Thus, compared to the transductive setting, where a model only has the ability to infer within trained components at seen timestamps, the extrapolation setting is more challenging to predict events, including unseen entities and relations at unseen timestamps. Some current embedding-based methods (e.g., MorsE~\citep{MorsE} and MaKEr~\citep{MaKEr}) have conducted extrapolation on static KGs, but they can not handle time information both in training TKGs and in test TKGs. Hence, we propose the problem of temporal link prediction with emerging entities and relations, called Temporal Knowledge Graph Extrapolation.\looseness=-1

\par
To address this issue, we resort to relative position patterns and temporal sequence patterns between relations to help our model understand the semantics of new relations and then pass these messages to new entities. Such patterns, which may imply causation and progressive relationships, are universal, entity-independent and transferable. Inspired by the recent successful applications of meta-learning models~\citep{metaleee}, which have the ability of “learning to learn”, in this paper, we propose a novel model, Meta-learning based Temporal Knowledge Graph Extrapolation (MTKGE). Our model contains three modules: 1) a Relative Position Pattern Graph (RPPG). In this module, we construct four kinds of relative position patterns between relations and reveal the structural features of relations in terms of space. 2) a Temporal Sequence Pattern Graph (TSPG), in which we learn three temporal sequence patterns between relations, reflecting the temporally structural features as an aspect of relation happening orders. 3) a GCN module for extrapolation. This module produces representations for both seen and unseen components to achieve Temporal Knowledge Graph Extrapolation.\looseness=-1

\par

\par
While meta-training, we sample a set of link prediction tasks from  existing (training) TKGs to simulate the scenario in emerging TKGs. Based on these tasks, we meta-train our model to obtain the embeddings of seen components and capture the patterns called meta-knowledge between relations stored in existing TKGs. 
MTKGE maps this meta-knowledge into an embedding space, and the feature of each seen relation is represented by fusing the embeddings of patterns with its connected relations so that a new emerging relation can also be represented.
We employ a graph neural network (GNN) to further incorporate temporal and position information into new entities according to their surrounding relations.
Hence, while extrapolating, the transferable meta-knowledge is leveraged to produce reasonable embeddings of unseen components in emerging TKGs. 
To verify the extrapolation ability of our proposed model MTKGE, we evaluate MTKGE using various score functions.
Experimental results show MTKGE outperforms the state-of-the-art models for knowledge graph extrapolation and specifically adapted KGE and TKGE baselines, due to the inclusion of time information and having the meta-learning ability, respectively. 
Compared to the strongest baseline results, MRR and Hits@10 of our model MTKGE increases by up to  70.2\% and 70.7\%.
\looseness=-1
\par
As far as we know, this work is the first attempt at knowledge extrapolation for TKGs using meta-learning based approach. In summary, the main contributions of this paper are as follows:\looseness=-1
\begin{itemize}[leftmargin=*]

\item The existing TKGE methods hold a closed-world assumption, where TKGs are fixed. We relax this assumption and propose a meta-learning based TKG extrapolation model, which can produce embeddings for emerging entities and relations, to predict missing links with unseen components in test TKGs.\looseness=-1

\item A relative position pattern graph module and a temporal sequence pattern graph module are proposed to capture transferable position and temporal meta-knowledge between relations.\looseness=-1

\item The experimental results on two TKG extrapolation benchmarks show that our model MTKGE  achieves the state-of-the-art on Temporal Knowledge Graph Extrapolation. Additionally, MTKGE shows remarkable extrapolation performance under various score functions and unseen ratios of entities and relations.\looseness=-1
\end{itemize}

\section{Related work}

\subsection{TKG Embedding }

Recently, there are several TKG Embedding models~\citep{10.5555/2999792.2999923,Dasgupta2018HyTEHT,jain-etal-2020-temporal, jiang-etal-2016-towards} encoding time information into low-dimension representations.

Following DistMult~\citep{DistMult}, TDistMult~\citep{TDistMult} learns one core
tensor for each timestamp based on Tucker decomposition.
RotatE~\citep{RotatE} views each relation as a rotation from the subject to the object and project entity embeddings from entity space to relation space, having the ability to infer various relation patterns.
Different from RotatE, TeRo~\citep{TeRo} defines the temporal evolution of entity embeddings as a rotation from the initial time to the current time in the complex vector space.
Following ComplEx~\citep{ComplEx}, TComplEx~\citep{TComplEx} considers time-dependent representation by adding a new time factor to extend the 3th-order tensor decomposition to 4th-order. ATiSE~\citep{ATISE} uses addictive time series decomposition to capture the evolution process of entities and relations by representing them as
Gaussian distribution.

However, these models can neither predict missing links in the unseen timestamps nor be directly used in the extrapolation setting.\looseness=-1

\subsection{TKG Extrapolation}
The Temporal Knowledge Graph Extrapolation problem, which this paper focuses on, aims to predict new events with unseen entities and relations based on seen facts. Graph Neural Network (GNN) shows an ability to encode non-linear architecture and features to help extrapolate. For static KG,  \citet{Hamaguchi2017KnowledgeTF} and \citet{Wang2019LogicAB}, based on neighbourhood aggregation, can only embed unseen entities connected with training KGs. SE-GNN~\citep{SEGNN} explore extrapolation problems from a data-relevant view, leveraging three kinds of neighbour patterns. GraiL~\citep{GraiL} and INDIGO~\citep{INDIGO} focus on inductive settings and extract subgraphs independent of any specific entities. However, such KG completion methods can not consider time information for both seen and unseen components. \looseness=-1
\par
In TKGs, the rule-based method Tlogic~\citep{Tlogic} makes use of random walking over TKGs to learn probabilistic logical rules, and such rules can be extrapolated to unseen timestamps. RE-GCN~\citep{REGCN} uses an auto-regressive approach and calculates along timestamps snapshots sequence capturing dynamic evolution based on GCN. In this work, we utilize the generalization ability of GNN to model rule-like patterns capturing the underlying logic between relations.\looseness=-1
\par
MaKEr~\citep{MaKEr} and MorsE~\citep{MorsE} are the most related works to ours, but they both deal with KG extrapolation in static graphs, neglecting the time information among events. 
Dynamic TKGs have temporally associated patterns between relations, which reflect the developing properties of real-world events. 
The temporal patterns, remaining constant for an extended period, can be captured by our meta-learning approach and transferred to unseen components. Motivated by this, we aim to explore the temporal patterns between relations into meta-knowledge and extrapolate them to emerging TKGs.\looseness=-1

\section{Problem Definition.} 
A temporal knowledge graph is defined as $ {G}=( {E},  {R},  {T},  {D})$, where $ {E}$ denotes a set of entities, $ {R}$ is a set of relations, $ {T}$ is a set of timestamps, and $ {D}$ is a set of quadruples. Specifically, $ {D}=\{(s, r, o, t)\} \subseteq  {E} \times  {R} \times  {E} \times  {T}$, where $s, o \in  {E}$, $r \in  {R}$ and $ t \in  {T}$.
To mimic the test environment, we formulate a set of tasks in training and test TKGs. For each task, we randomly extract a sub-TKG and treat entities and relations between them as unseen. Test TKGs are divided into support quadruples $ {D}_{\text {sup}}^{t e}$ and query quadruples $ {D}_{\text {que}}^{t e}$. The support quadruples $ {D}_{\text {sup}}^{t e}$  reveal the interaction of entities and relations with timestamps, including new components. Then these connecting relationships are used to produce embeddings of both seen and unseen entities. The query quadruples $ {D}_{\text {que}}^{t e}$ in test TKGs are used for testing the performance of the model by evaluating the rationality of these quadruples based on produced embeddings.\looseness=-1
\par
Based on the above definitions, we formally define the proposed problem of Temporal Knowledge Graph Extrapolation in our paper. Given a training TKG $ {G}^{t r}$=($ {E}^{t r}$, $ {R}^{t r}$, $ {T}^{t r}$ , $ {D}^{t r}$) and a test TKG $ {G}^{t e}$=$ {E}^{t e}$, $ {R}^{t e}$, $ {T}^{t e}$,  $ {{D}_{ s u p}}^{t e}$, $ {{D}_{ q u e}}^{t e})$, where $ {E}^{t r}$ $\neq$ $ {E}^{t e}$, $ {E}^{t r}$ $\cap$ $ {E}^{t e}$ $\neq$ $\emptyset$, and $ {R}^{t r}$ $\neq$ $ {R}^{t e}$, $ {R}^{t r}$ $\cap$ $ {R}^{t e}$ $\neq$ $\emptyset$, and $ {T}^{t r}$ $\neq$ $ {T}^{t e}$, $ {T}^{t r}$ $\cap$ $ {T}^{t e}$ $\neq$ $\emptyset$, the problem asks to train a temporal link prediction model based on tasks sampled from  $ {D}^{t r}$ in $ {G}^{t r}$ and $ {{D}_{ s u p}}^{t e}$ in $ {G}^{t e}$; the goal of the problem is to predict missing links with 
emerging entities and relations of $ {{D}_{ q u e}}^{t e}$ in $ {G}^{t e}$. A sampled task $ {S}^{i}$ is defined as:
\begin{equation}
\begin{split}
 {S}^{i} = \Big(  & {E}^{i} = \left( \widehat{ {E}}^{i} ,  \widetilde{ {E}}^{i} \right) ,   {R}^{i} = \left( \widehat{ {R}}^{i} , \widetilde{ {R}}^{i}  \right),
\\
&  {T}^{i} = \left( \widehat{ {T}}^{i} ,  \widetilde{ {T}}^{i} \right) ,   {D}_{s u p}^{i} ,   {D}_{q u e}^{i} \Big),
\end{split}
\label{equ1}
\end{equation}
where \;$\widehat{}$ \;and \;$\widetilde{}$ \;stand for seen components and emerging components respectively; $ {D}_{s u p}^{i} \subseteq  {D}_{s u p}^{t e}$; $ {D}_{q u e}^{i} \subseteq  {D}_{q u e}^{t e}$.
\begin{figure*}[!]
    \centering

    \resizebox{50em}{!}{\includegraphics{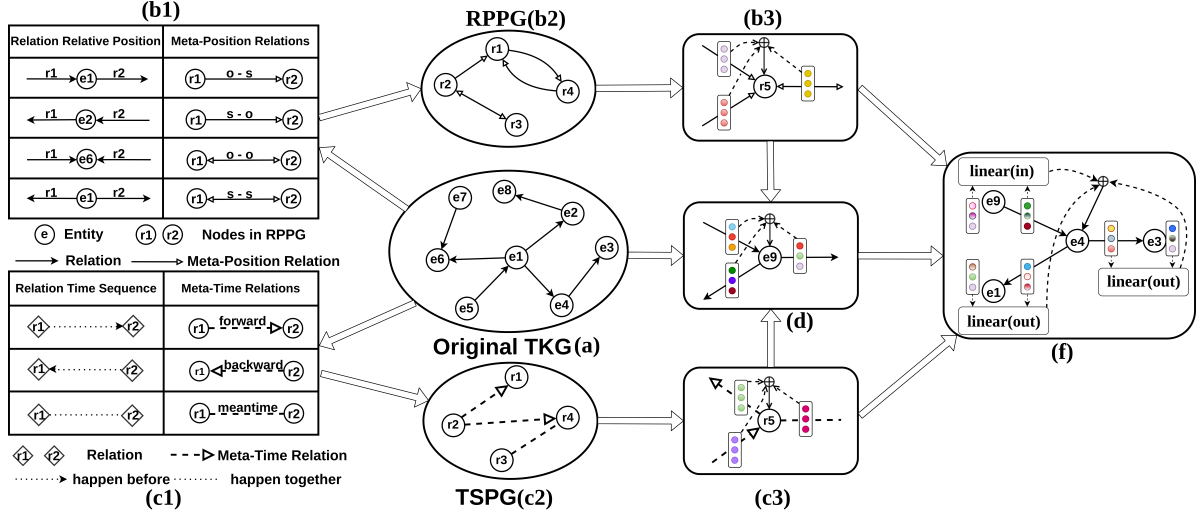}}
    \caption{Overview of our proposed model MTKGE. Starting from (a), (b1)(b2)(b3) construct RPPG from (a); (c1)(c2)(c3) construct TSPG from (a); (d) outputs unseen entity embeddings; (f) conducts extrapolation.}
    \label{mainfig}
    \vspace{-0.2cm}
\end{figure*}
\section{Proposed Approach}
\subsection{Overview}
Our model is mainly based on graph neural network (GNN)  and follows the widely used encoder-decoder framework~\citep{encoderdecoder}. The encoder leverages GNN, which can encode non-linearity architecture information, to mine abundant meta-knowledge hidden in training TKGs. The decoder, one of six KGE or TKGE models used as a score function, takes the embeddings of the subject, relation, object and timestamp in a quadruple as input and calculates them into a score reflecting the rationality of the quadruple. \looseness=-1
\par
Fig.~\ref{mainfig} depicts the architecture of our model MTKGE. We illustrate the overall process on a single task $ {S}^{i}$ as defined in (\ref{equ1}). Naturally, the connected relations to a seen entity indicate their entity-related features. We argue relations themselves also have entity-independent characteristics, which reflect the logical patterns between relations. 

However, as mentioned above, the main challenge in Temporal Knowledge Graph Extrapolation is to produce embeddings for unseen entities and relations. To tackle this problem, we design two GNN modules to capture important entity-independent relation features: the Relative Position Pattern Graph (RPPG) and the Temporal Sequence Pattern Graph (TSPG). Then, these pattern features, called meta-knowledge, are aggregated in the Entity Feature Representation module and the Temporal Knowledge Extrapolation module to generate reasonable embeddings for unseen components.\looseness=-1
\subsection{Relative Position Pattern Graph}
\par
Inspired by the transferable learning for GNN proposed by \citep{transferGNN}, where node features like node degrees and k-hop expansion subgraphs are considered transferable in terms of structure, in our work, we note that relations themselves also have structure-respecting features because of homophily among graphs. Specifically, there are four reusable position patterns between relations, as shown in Fig.~\ref{mainfig}(b1). We refer to them as meta-position relations.\looseness=-1
\par
In order to extract these meta-position relations, we construct a Relative Position Patterns Graph (RPPG, Fig.~\ref{mainfig}(b2)), in which we regard relations in the original TKG (Fig.~\ref{mainfig}(a)) as nodes. The meta-position relations we defined connect nodes in RPPG and reflect the relative position features of relations in the original TKG. For example, if a entity serves as object of relation $r_{1} $ and serves as subject of relation $r_{2} $ regardless of time, we add an edge $\left(r_{1}, s-o, r_{2}\right)$ to RPPG. After building the RPPG from the original TKG, as shown in Fig.~\ref{mainfig}(b3), we can easily obtain the first part feature of an unseen relation \emph{r} represented as $\mathbf{g}_{r}$, with an aggregation which combines its neighbour meta-position relations in RPPG :\looseness=-1
\begin{equation}
\mathbf{g}_{r}=\frac{1}{| {N}_{p}(r)|} \sum_{m \in  {N}_{p}(r)} \mathbf{g}_{m},
\end{equation}
where  $ {N}_{p}(r)$ denotes the set of in-going meta-position relations of the unseen relation \emph{r}. $\mathbf{g}_{m} \in  {G}_{M}$  represents the embeddings of the meta-position relation type, and $ {G}_{M}=\left(\mathbf{g}_{\mathrm{o}-\mathrm{s}}, \mathbf{g}_{\mathrm{s}-\mathrm{o}}, \mathbf{g}_{\mathrm{o}-\mathrm{o}}, \mathbf{g}_{\mathrm{s}-\mathrm{s}}\right)$ parameterizes the four types of meta-position relations.\looseness=-1
\subsection{Temporal Sequence Pattern Graph}
\par
Apart from relative position patterns of relations, temporal sequence patterns naturally exist between relations. As shown in Fig.~\ref{fig1}, $ {engage}$ $ {in}$ $ {cooperation}$ occurred after $ {express}$ $ {intent}$ $ {to}$ $ {negotiate}$ and $ {make}$/$ {host}$ $ {a}$ $ {visit}$; $ {express}$ $ {intent}$ $ {to}$ $ {negotiate}$ happened between $ {accuse}$ and $ {make}$/$ {host}$ $ {a}$ $ {visit}$. 
To model such relations orders in time, we design three kinds of meta-time relations (Fig.~\ref{mainfig}(c1)) and construct a Temporal Sequence Patterns Graph (TSPG, Fig.~\ref{mainfig}(c2)). For instance, $\left(r_{1}, forward, r_{2}\right)$ in TSPG denotes relation $r_{2} $ happens after another relation $r_{1} $. 
We calculate the second part feature of an unseen relation \emph{r} represented as $\mathbf{g}_{r}$ as follows, as shown in Fig.~\ref{mainfig}(c3):
\begin{equation}
\mathbf{q}_{r}=\frac{1}{| {N}_{t}(r)|} \sum_{n \in  {N}_{t}(r)} \mathbf{q}_{n},
\end{equation}
where $ {N}_{t}(r)$ denotes the set of three types of meta-time relations of \emph{r}, which are parameterized by $ {Q}_{N}=\left( \mathbf{q}_{\mathrm{forward}}, \mathbf{q}_{\mathrm{backward}}, \mathbf{q}_{\mathrm{meantime}} \right)$. Finally, features of the unseen relation \emph{r} is denoted as $\mathbf{h}_{r}$ = [$\mathbf{g}_{r}$; $\mathbf{q}_{r}$] by concatenating the two parts above. 

\subsection{Entity Feature Representation} 
\par
All the existing and emerging relations have been explored in the above two modules. 
For unseen entities, we use their surrounding relations in the original TKG to represent their embeddings. The transferred meta-position relation features and meta-time relation features enrich unseen entities with reasonable semantic characters. We consider the direction of relations connected to the unseen entity \emph{e} represented as $\mathbf{h}_{e}$, as shown in Fig.~\ref{mainfig}(d):
\begin{equation}
\mathbf{h}_{e}=\frac{1}{| {N}(e)|} \sum_{r \in  {N}(e)} \mathbf{W}_{\operatorname{dir}(r)}^{\mathrm{ent}} \mathbf{h}_{r},
\end{equation}
where $ {N}(e)$ is a set of relations connected to \emph{e}, dividing into in-going and out-going. Therefore,  $\mathbf{W}_{\operatorname{dir}(r)}^{\text {ent }}$, which is a direction-specific parameter, divides into $\left \{   \mathbf{W}_{\text {in }}^{\text {ent }},  \mathbf{W}_{\text {out }}^{\text {ent }}  \right \}$. \looseness=-1
\begin{table*}[t]

\centering

\resizebox{\linewidth}{!}{
\begin{tabular}{|c|c|c|c|c|c|c|c|c|c|c|c|c|c|c|}
\specialrule{0.2em}{1pt}{1pt}
\multicolumn{1}{c}{}& \multicolumn{4}{c}{Training TKG}&\multicolumn{5}{c}{Test TKG}&\multicolumn{5}{c}{Valid TKG}\\
\cmidrule(r){2-5}  \cmidrule(r){6-10}  \cmidrule(r){11-15}
\multicolumn{1}{c}{}& \multicolumn{1}{c}{$\left| {E}^{t r}\right|$}& \multicolumn{1}{c}{$\left| {R}^{t r}\right|$}& \multicolumn{1}{c}{$\left| {T}^{t r}\right|$}& \multicolumn{1}{c}{$\left| {D}^{t r}\right|$}& \multicolumn{1}{c}{$\left| {E}^{t e}\right|$}& \multicolumn{1}{c}{$\left| {R}^{t e}\right|$}& \multicolumn{1}{c}{$\left| {T}^{t e}\right|$}& \multicolumn{1}{c}{$\left| {D}_{\text {sup }}^{t e}\right|$}& \multicolumn{1}{c}{$\left| {D}_{\text {que }}^{t e}\right|$}& \multicolumn{1}{c}{$\left| {E}^{v a}\right|$}& \multicolumn{1}{c}{$\left| {R}^{v a}\right|$}& \multicolumn{1}{c}{$\left| {T}^{v a}\right|$}& \multicolumn{1}{c}{$\left| {D}_{\text {sup }}^{v a}\right|$}& \multicolumn{1}{c}{$\left| {D}_{\text {sup }}^{v a}\right|$}
\\[0.8ex]
\specialrule{0.1em}{1pt}{1pt}

\multicolumn{1}{c}{ IC14-Ext} & \multicolumn{1}{c}{502} & \multicolumn{1}{c}{96} & \multicolumn{1}{c}{291}  & \multicolumn{1}{c|}{3017} & \multicolumn{1}{c}{574(495)} & \multicolumn{1}{c}{204(112)} &\multicolumn{1}{c}{365(74)}  &\multicolumn{1}{c}{25827} &\multicolumn{1}{c|}{7477}  &\multicolumn{1}{c}{503(408)}  &\multicolumn{1}{c}{131(46)} & \multicolumn{1}{c}{292(1)} & \multicolumn{1}{c}{5998} & \multicolumn{1}{c}{713}
 \\[0.8ex]
 
\multicolumn{1}{c}{ IC0515-Ext} & \multicolumn{1}{c}{465} & \multicolumn{1}{c}{143} & \multicolumn{1}{c}{2857}  & \multicolumn{1}{c|}{11304} & \multicolumn{1}{c}{550(470)} & \multicolumn{1}{c}{236(95)} &\multicolumn{1}{c}{4017(1160)}  &\multicolumn{1}{c}{162558} &\multicolumn{1}{c|}{56367}  &\multicolumn{1}{c}{555(455)}  &\multicolumn{1}{c}{167(33)} & \multicolumn{1}{c}{3171(340)} & \multicolumn{1}{c}{27688} & \multicolumn{1}{c}{3198}
 \\[0.8ex]
 
\specialrule{0.2em}{1pt}{1pt}
\end{tabular}
}
\caption{\label{table2}Statistics of our datasets IC14-Ext and IC0515-Ext. The numbers in the bracket denote the number of components that do not appear in the corresponding training TKGs (i.e., unseen components).}
\vspace{-0.6cm}
\end{table*}

\subsection{Temporal Knowledge Extrapolation} 
\par
Based on CompGCN~\citep{COMPGCN}, we propose a GCN module to aggregate k-hop relation information. While extrapolating, there are two types of components: seen and unseen. The embeddings of unseen relations are produced by RPPG and TSPG, and the Entity Feature Representation module produces the embeddings of unseen entities, while the seen components should directly get from learned feature matrices during training. The feature aggregation of an entity \emph{e} at $\ell+1$-th layer of GCN represented as $\mathbf{m}_{e}^{\ell+1}$ is calculated as:\looseness=-1
\begin{equation}
\begin{aligned}
\setlength{\abovedisplayskip}{-8pt}
\setlength{\belowdisplayskip}{-10pt}
\begin{split}
\mathbf{m}_{e}^{\ell+1}=&\sum_{(r, o) \in  {O}(e)} \quad \mathbf{W}_{\text {out }}^{\ell}\left[\mathbf{h}_{r}^{\ell} ; \mathbf{h}_{o}^{\ell};\mathbf{h}_{t}^{\ell}\right]+\\
&\sum_{( s,r) \in  {I}(e)} \quad \mathbf{W}_{\text {in }}^{\ell}  \left[\mathbf{h}_{r}^{\ell} ; \mathbf{h}_{s}^{\ell};\mathbf{h}_{t}^{\ell}\right],
\end{split}
\end{aligned}
\end{equation}
where `;' is the concatenation operation, $ {O}(e)$ and $ {I}(e)$ denote a set of out-going and in-going relations and their  connecting entities. $\mathbf{W}_{\text {out }}^{\ell}$ and $\mathbf{W}_{\text {in }}^{\ell}$ are learnable parameters for relation-entity pairs at the $\ell$-th layer of GCN. We initialize $\mathbf{h}_{r}^{0}$ , $\mathbf{h}_{s}^{0}$ and $\mathbf{h}_{o}^{0}$ as $\mathbf{h}_{r}$ , $\mathbf{h}_{s}$ and $\mathbf{h}_{o}$, which are input relation and entity features, and we randomly initialize $\mathbf{h}_{t}$. In Fig.~\ref{mainfig}(d), after k-hop updating for each entity, the entity representation for each \emph{e}, $\mathbf{h}_{e}^{k+1}$, is calculated by:
\begin{equation}
\mathbf{h}_{e}^{k+1}=\sigma\left(\frac{\mathbf{m}_{e}^{k+1}}{| {O}(e)|+| {I}(e)|}+\mathbf{W}_{\text {self }}^{k} \mathbf{h}_{e}^{k}\right),
\end{equation}
where $\mathbf{W}_{\text {self }}^{k}$ is a learnable parameter for self-loop updating for entities, and $\sigma$ is an activation function. Relations and timestamps are also updated at each layer as follows:
\begin{equation}
\mathbf{h}_{r}^{\ell+1}=\sigma\left( \mathbf{W}_{\text {rel }}^{\ell} \mathbf{h}_{r}^{\ell} \right) ;   \mathbf{h}_{t}^{\ell+1}=\sigma\left( \mathbf{W}_{\text {time }}^{\ell} \mathbf{h}_{t}^{\ell} \right).
\end{equation}
\par
The final outputs of entities and relations representations are exactly the embeddings for both seen and unseen entities and relations, achieving Temporal Knowledge Graph Extrapolation.

\subsection{Model Learning}
\par
For a task $ {S}^{i}$, we meta-train our model in the training TKG to gain meta-knowledge,  extrapolate in the support quadruples $ {D}_{\text {sup }}^{i}$ in the test TKG and finally test in the query quadruples $ {D}_{\text {que }}^{i}$ in the test TKG. Moreover, the rationality evaluation of quadruples is based on the score functions in various KGE models (ComplEx, DistMult, RotatE) and their corresponding TKGE versions (TComplEx, TDistMult, TeRo). We apply the self-adversarial negative sampling loss function proposed by \citep{RotatE} on query quadruples $ {D}_{\text {que }}^{i}$ :
{
\begin{equation}
\begin{aligned}
\begin{split}
 {L}\left( {S}^{i}\right)&= \frac{1}{\left| {D}_{q u e}^{i}\right|} \sum_{(s, r, o, t) \in  {D}_{\text {que }}^{i}}-\log \sigma(\gamma+\varphi (s, r, o, t)) \\
&-\sum_{i=1}^{n} p\left(s_{i}^{\prime}, r, o_{i}^{\prime}, t\right) \log \sigma\left(-\gamma-\varphi \left(s_{i}^{\prime}, r, o_{i}^{\prime}, t\right)\right),
\end{split}
\label{rotateloss}
\end{aligned}
\end{equation}
}
where $\varphi (s, r, o, t)$ is the score function for (\emph{s}, \emph{r}, \emph{o}, \emph{t}), $\gamma$ is a fixed margin, $\emph{n}$ is the number of negative samples, $\left(s_{i}^{\prime}, r, o_{i}^{\prime}, t\right)$ is a negative sample by corrupting a subject or object entity. $p\left(s_{i}^{\prime}, r, o_{i}^{\prime}, t\right)$ is the weight for a negative sample. we put its calculation in Appendix \ref{loss}. Note that while using KGE score functions, the time factor is ignored.  Finally, the overall loss among all tasks sampled 
is  $\sum_{i}  {L}\left( {S}^{i}\right)$.\looseness=-1

\section{Experiment}

\setlength{\parskip}{0.1cm}

\subsection{Datasets}

\par
In traditional TKG datasets, such as ICEWS14, ICEWS05-15, YAGO11K and Wikidata12k, the sets of entities and relations in test quadruples belong to the sets of entities and relations in train quadruples. Therefore, the two datasets are unsuitable for temporal knowledge graph completion in the extrapolation setting.
ICEWS has more relation types than YAGO11K and Wikidata12k, resulting in richer and more challenging relation patterns. Also, our extrapolation modules capture the intrinsic temporal development order of relations in the real world without introducing domain-specific information. Therefore, aiming to evaluate the transferability of meta-knowledge learned by our model, we build two new datasets from ICEWS14 and ICEWS05-15, named IC14-Ext and IC0515-Ext. \looseness=-1
\par
For each dataset, we sample a test TKG and its corresponding training TKG from the original benchmark. Carrying out masking at a random ratio ensures unseen entities and relations exist in test TKGs, where query quadruples contain at least one unseen component. They can be further classified into quadruples only containing unseen entities ($\text {\emph{u}\_\emph{ent}}$), only containing unseen relations ($\text {\emph{u}\_\emph{rel}}$), and containing both unseen entities and unseen relations($\text {\emph{u}\_\emph{both}}$). The statistics of the two datasets are given in Table~\ref{table2}. The numbers of query quadruples for $\text {\emph{u\_\emph{ent}}}$,$\text {\emph{u}\_\emph{rel}}$ and $\text {\emph{u}\_\emph{both}}$ are 3486, 2059, 1932 in IC14-Ext, and 22487, 17331, 16549 in IC0515-Ext. The details of generating our datasets are in Appendix~\ref{A}.\looseness=-1

\begin{table*}[t]
\centering
\resizebox{\linewidth}{!}{
\begin{tabular}{|c|c|c|c|c|c|c|c|c|c|c|c|c|}
\specialrule{0.2em}{1pt}{1pt}
\multicolumn{1}{c}{}& \multicolumn{6}{c}{IC14-Ext}&\multicolumn{6}{c}{IC0515-Ext}\\
\cmidrule(r){2-7}  \cmidrule(r){8-13}
\multicolumn{1}{c}{}& \multicolumn{2}{c}{$\text{\emph{u}\_\emph{ent}}$}&
\multicolumn{2}{c}{$\text {\emph{u}\_\emph{rel}}$}&
\multicolumn{2}{c}{$\text {\emph{u}\_\emph{both}}$}& \multicolumn{2}{c}{$\text{\emph{u}\_\emph{ent}}$}&
\multicolumn{2}{c}{$\text {\emph{u}\_\emph{rel}}$}&
\multicolumn{2}{c}{$\text {\emph{u}\_\emph{both}}$}
\\
  \cmidrule(r){2-3}  \cmidrule(r){4-5} \cmidrule(r){6-7}\cmidrule(r){8-9}\cmidrule(r){10-11}\cmidrule(r){12-13}
\multicolumn{1}{c}{} & \multicolumn{1}{c}{MRR} & \multicolumn{1}{c}{Hits@10} & \multicolumn{1}{c}{MRR} & \multicolumn{1}{c}{Hits@10}  & \multicolumn{1}{c}{MRR} & \multicolumn{1}{c}{Hits@10}
 & \multicolumn{1}{c}{MRR} & \multicolumn{1}{c}{Hits@10} & \multicolumn{1}{c}{MRR} & \multicolumn{1}{c}{Hits@10} & \multicolumn{1}{c}{MRR} & \multicolumn{1}{c}{Hits@10}
 \\
\specialrule{0.1em}{1pt}{1pt}
\multicolumn{1}{c}{ Asmp-TKGE(TComplEx)} & \multicolumn{1}{c}{0.0914} & \multicolumn{1}{c}{19.67} & \multicolumn{1}{c}{0.1052}  & \multicolumn{1}{c}{20.11} & \multicolumn{1}{c}{0.1188} & \multicolumn{1}{c|}{20.80} &\multicolumn{1}{c}{0.1065}  &\multicolumn{1}{c}{20.18} &\multicolumn{1}{c}{0.1105}  &\multicolumn{1}{c}{22.69}  &\multicolumn{1}{c}{0.0988} & \multicolumn{1}{c}{14.65}
 \\[0.8ex]
\multicolumn{1}{c}{ Asmp-TKGE(TDistMult)} & \multicolumn{1}{c}{0.1565} & \multicolumn{1}{c}{23.64} & \multicolumn{1}{c}{\underline{0.1786}}  & \multicolumn{1}{c}{25.75} & \multicolumn{1}{c}{0.1688} & \multicolumn{1}{c|}{25.66} &\multicolumn{1}{c}{0.1923}  &\multicolumn{1}{c}{20.11} &\multicolumn{1}{c}{0.1856}  &\multicolumn{1}{c}{19.84}  &\multicolumn{1}{c}{0.1805} & \multicolumn{1}{c}{20.51}
 \\[0.8ex]

 \multicolumn{1}{c}{ Asmp-TKGE(TeRo)} & \multicolumn{1}{c}{0.1894} & \multicolumn{1}{c}{20.31} & \multicolumn{1}{c}{0.1633}  & \multicolumn{1}{c}{18.39} & \multicolumn{1}{c}{0.1742} & \multicolumn{1}{c|}{20.97} &\multicolumn{1}{c}{0.1555}  &\multicolumn{1}{c}{17.88} &\multicolumn{1}{c}{0.1899}  &\multicolumn{1}{c}{19.23}  &\multicolumn{1}{c}{0.1782} & \multicolumn{1}{c}{19.01}
 \\[0.8ex]
 \specialrule{0.1em}{1pt}{1pt}
 \multicolumn{1}{c}{ Asmp-KGE(ComplEx)} & \multicolumn{1}{c}{0.1102} & \multicolumn{1}{c}{22.44} & \multicolumn{1}{c}{0.1077}  & \multicolumn{1}{c}{21.69} & \multicolumn{1}{c}{0.1256} & \multicolumn{1}{c|}{21.62} &\multicolumn{1}{c}{0.1294}  &\multicolumn{1}{c}{22.99} &\multicolumn{1}{c}{0.1025}  &\multicolumn{1}{c}{20.09}  &\multicolumn{1}{c}{0.0906} & \multicolumn{1}{c}{13.33}
 \\[0.8ex]
\multicolumn{1}{c}{ Asmp-KGE(DistMult)} & \multicolumn{1}{c}{0.1322} & \multicolumn{1}{c}{25.14} & \multicolumn{1}{c}{0.1399}  & \multicolumn{1}{c}{24.76} & \multicolumn{1}{c}{0.1966} & \multicolumn{1}{c|}{28.60} &\multicolumn{1}{c}{0.2465}  &\multicolumn{1}{c}{26.11} &\multicolumn{1}{c}{0.1890}  &\multicolumn{1}{c}{17.84}  &\multicolumn{1}{c}{0.1895} & \multicolumn{1}{c}{21.59}
 \\[0.8ex]

 \multicolumn{1}{c}{ Asmp-KGE(RotatE)} & \multicolumn{1}{c}{0.1962} & \multicolumn{1}{c}{21.39} & \multicolumn{1}{c}{0.1785}  & \multicolumn{1}{c}{19.88} & \multicolumn{1}{c}{\underline{0.1796}} & \multicolumn{1}{c|}{21.23} &\multicolumn{1}{c}{0.1962}  &\multicolumn{1}{c}{19.51} &\multicolumn{1}{c}{0.1766}  &\multicolumn{1}{c}{18.29}  &\multicolumn{1}{c}{0.1932} & \multicolumn{1}{c}{22.65}
 \\[0.8ex]
 \specialrule{0.1em}{1pt}{1pt}
\multicolumn{1}{c}{SE-GNN} & \multicolumn{1}{c}{0.1076} & \multicolumn{1}{c}{19.10} & \multicolumn{1}{c}{0.1354}  & \multicolumn{1}{c}{24.40} & \multicolumn{1}{c}{0.1115} & \multicolumn{1}{c|}{18.70} &\multicolumn{1}{c}{0.1246}  &\multicolumn{1}{c}{24.10} &\multicolumn{1}{c}{0.1952}  &\multicolumn{1}{c}{22.58}  &\multicolumn{1}{c}{0.1203} & \multicolumn{1}{c}{18.20}
 \\[0.8ex]
\multicolumn{1}{c}{ MaKEr} & \multicolumn{1}{c}{\underline{0.2562}} & \multicolumn{1}{c}{\underline{47.87}} & \multicolumn{1}{c}{0.1435}  & \multicolumn{1}{c}{\underline{31.00}} & \multicolumn{1}{c}{0.1512} & \multicolumn{1}{c|}{\underline{32.00}} &\multicolumn{1}{c}{\underline{0.3489} } &\multicolumn{1}{c}{\underline{52.42}} &\multicolumn{1}{c}{\underline{0.2790}}  &\multicolumn{1}{c}{\underline{48.73}}  &\multicolumn{1}{c}{\underline{0.3536}} & \multicolumn{1}{c}{\underline{55.58}}
 \\[0.8ex]

\specialrule{0.1em}{1pt}{1pt}

\multicolumn{1}{c}{MTKGE} & \multicolumn{1}{c}{\textbf{0.3268}} & \multicolumn{1}{c}{\textbf{53.41}} & \multicolumn{1}{c}{\textbf{0.2817}}  & \multicolumn{1}{c}{\textbf{52.91}} & \multicolumn{1}{c}{\textbf{0.3056}} & \multicolumn{1}{c|}{\textbf{49.27}} &\multicolumn{1}{c}{\textbf{0.4023}}  &\multicolumn{1}{c}{\textbf{60.88}} &\multicolumn{1}{c}{\textbf{0.3016}} &\multicolumn{1}{c}{\textbf{61.08}}  &\multicolumn{1}{c}{\textbf{0.4104}} & \multicolumn{1}{c}{\textbf{66.88}}

\\[0.8ex]
\multicolumn{1}{c}{$ {Improv.}$} & \multicolumn{1}{c}{27.6\%} & \multicolumn{1}{c}{11.6\%} & \multicolumn{1}{c}{57.7\%}  & \multicolumn{1}{c}{70.7\%} & \multicolumn{1}{c}{70.2\%} & \multicolumn{1}{c|}{54.0\%} &\multicolumn{1}{c}{15.3\%}  &\multicolumn{1}{c}{16.1\%} &\multicolumn{1}{c}{8.1\%}  &\multicolumn{1}{c}{25.3\%}  &\multicolumn{1}{c}{16.1\%} & \multicolumn{1}{c}{20.3\%}

\\
\specialrule{0.2em}{1pt}{1pt}

\end{tabular}}

\caption{\label{table1}Link prediction results (\% for Hits@10) on two datasets. Asmp-KGE(RotatE) means specifically adapted KGE method RotatE, the same for other Asmp baselines. We show results for query quadruples only containing unseen entities ($\text{\emph{u}\_\emph{ent}}$), only containing unseen relations ($\text{\emph{u}\_\emph{rel}}$) and containing both unseen entities and relations ($\text{\emph{u}\_\emph{both}}$). \textbf{Bold} numbers denote the best results and \underline{underline} numbers denote the strongest baseline results in all kinds of methods. }
\vspace{-0.6cm}
\end{table*}

\subsection{Baselines} 

We compare our model with 1)MaKEr~\citep{MaKEr} and SE-GNN~\citep{SEGNN}, the state-of-the-art models of knowledge graph completion in the extrapolation setting; 2) three specifically adapted TKGE baselines, including TComplEx~\citep{TComplEx}, TDistMult~\citep{TDistMult}, TeRo~\citep{TeRo}; and 3) three specifically adapted KGE baselines, including ComplEx~\citep{ComplEx}, DistMult~\citep{DistMult}, RotatE~\citep{RotatE}.
SE-GNN is a GNN-based model depending on three-fold extrapolative knowledge representation and uses ConvE ~\citep{Dettmers2018Convolutional2K} as a decoder, which uses 2D convolutional neural networks to match queries and answers.

Even though MaKEr and SE-GNN do not consider time information, the two models are powerful GNN-based representative baselines in the extrapolation setting. \looseness=-1
\par
For TKGE baselines, they are widely proven to be effective in the transductive setting~\citep{Wang2014KnowledgeGE,Lin2015LearningEA, Feng2016KnowledgeGE, balazevic-etal-2019-tucker,Kazemi2018SimplEEF,Qian2018TranslatingEF}, where all components tested are well-trained within fixed TKGs with no new entities or relations. In order to adapt the original models to fit our extrapolation setting, we follow their principle to embed emerging entities and relations. 
For example, TeRo~\citep{TeRo} regards relations as rotation from head entities to tail entities, and its score function is $f_{\mathrm{TeRo}}(s, r, o, t)=\left\|\emph{s} \circ \emph{t}+\emph{r}-\overline{\emph{o} \circ \emph{t}}\right\|$, where $\circ$ denotes the Hermitian dot product between complex vectors.
Specifically, we first train a conventional TeRo in the training TKG obtaining the embeddings of seen components and then leverage the corresponding core modelling way to calculate embeddings for unseen entities and relations. 
For instance,  a quadruple $\left(s, r, o, t\right)$ in the support quadruples $ {D}_{\text {sup }}^{t e}$ of test TKG where the head entity $s$, the relation $r$ and the timestamp $t$ are seen, the embedding of the emerging tail entity $o$  can be calculated by $\emph{o} = \overline{(\emph{s} \circ \emph{t} + \emph{r})} / \emph{t}, $ which leads the score function $f_{\mathrm{TeRo}}(s, r, o, t)$ to 0 and matches the principle of TeRo. Well-adapted calculating details for other TKGE and KGE baselines in the TKG Extrapolation problem are given in Appendix~\ref{E}. We refer to them as Asmp-TKGE and Asmp-KGE in our paper. \looseness=-1

\subsection{Implementation Details } 
\par
We report Mean Reciprocal Rank (MRR) and Hits at 1, 10 (Hits@1, @10) to evaluate the link prediction performance of query quadruples in the test TKG for each dataset. The link prediction includes three situations (i.e., $\text {\emph{u}\_\emph{ent}}$, $\text {\emph{u}\_\emph{rel}}$, and $\text {\emph{u}\_\emph{both}}$) and considers both head and tail link prediction tasks. 
\par
Our model is implemented in PyTorch and DGL, and all experiments are carried out on Tesla V100. 
For MaKEr, we use the implementations publicly provided by the authors with their best configurations. MaKEr can only deal with unseen entities and relations regardless of time, so we directly skip the time factor while running MaKEr on IC14-Ext and IC0515-Ext. For Asmp-TKGE and Asmp-KGE, we specifically adapted the public code\!~\footnote{\url{https://github.com/facebookresearch/tkbc} for TKGE: TComplEx, TDistMult and TeRo;\\ \url{https://github.com/mniepert/mmkb} for KGE: ComplEx, DistMult and RotatE.} and the embedding dimension is 128. For MTKGE, the dimension for embeddings and feature representations is 128; we employ the GNN with two layers, and the dimension for GNN’s hidden representation is 64. The batch size for meta-training is 64, and we use the Adam optimizer with a learning rate in a range of \{0.0001, 0.001, 0.01, 0.1\}. The time regularization parameter, denoted as reg, is tuned in \{0, 0.0001, 0.001, 0.01 , 0.1\}. Before meta-training our model, we sample 10,000 link prediction tasks on the existing TKGs for each dataset. To evaluate the average performance of our model, we randomly generate the ratios of unseen entities and relations with a range of 30$\%$ \textasciitilde 70$\%$ while sampling each task. For further research, we conduct a robustness study to analyze the effect of the ratios of unseen entities and relations to the final results. The details can be found in section \ref{Robustness Study}.\looseness=-1

\subsection{Experimental Results and Analysis}
\par
Table~\ref{table1} shows the link prediction results for three different kinds of query quadruples (i.e., $\text {\emph{u}\_\emph{ent}}$, $\text {\emph{u}\_\emph{rel}}$ and $\text {\emph{u}\_\emph{both}}$) on datasets IC14-Ext and IC0515-Ext.
Note that both our model and the baseline MaKEr are embedding-based frameworks that can be equipped with different decoders. In order to find the most suitable KGE or TKGE methods as the decoder, results of MTKGE and MaKEr equipped with three typical KGE score functions, ComplEx, DistMult, RotatE, and three corresponding temporal variants TKGE score functions, TComplEx, TDistMult, TeRo, are considered. The best results among all the decoders reflect the actual model performance, so in Table~\ref{table1}, we report the best results of MaKEr and MTKGE. 
For further comparison, in Figure~\ref{comparision} in Appendix~\ref{D}, we show all variants of our model MTKGE using the above six score functions and make a fair and transparent comparison with MaKEr \citep{MaKEr} using the same score function, respectively. \looseness=-1


\subsubsection{\textbf{Experimental Results}}

From Table~\ref{table1}, we can see that our proposed model MTKGE 
remarkably outperforms all state-of-the-art baselines and specifically adapted Asmp baselines on the two datasets across all metrics consistently. 
Among three kinds of query quadruples in IC14-Ext, compared with Asmp-KGE and Asmp-TKGE, the MRR results of MTKGE averagely increase by 14\% (\emph{u}\_\emph{ent}), 58\% (\emph{u}\_\emph{rel}), 87\% (\emph{u}\_\emph{both}) relatively. 
The results of Asmp baselines mean that although these adapted KGE and TKGE baselines have a particular ability to extrapolate emerging components, there is still a bias to conduct convincing extrapolation for them. 
Our model also notably outperforms MaKEr. More precisely, MTKGE averagely increases by 0.75 and 0.45 times for MRR and Hits@10 on IC14-Ext dataset and 0.13 and 0.16 times on IC0515-Ext dataset.
The $ {Improv.}$ in Table~\ref{table1} shows the improvement percentage of our model against the strongest baseline(underline numbers). On both datasets, the highest improvement percentages for MRR and Hits@10 are 70.2\% and 70.7\%, proving that our proposed MTKGE gains significant enhancement on Temporal Knowledge Graph Extrapolation.\looseness=-1

\par
From Figure~\ref{comparision}, we can see that our proposed model MTKGE has a higher performance in all cases compared to MaKEr with the same score function. Hence, MTKGE could leverage these six score functions better than MaKEr does. 
Figure~\ref{comparision} also shows that both KGE and TKGE serving as score functions have a significant influence on the performance of both models, which is because different score functions judge the rationality of the quadruplets from various perspectives. 
Despite this, some score functions still have considerable benefits to the encoding models(MTKGE and MaKEr). Specifically, MTKGE using
tensor decomposition-based TKG embedding models, 
such as MTKGE(TDistMult) and MTKGE (TComplEx), have comparable performance to the red line in Figure~\ref{comparision}, which is the best performance achieved by MTKGE(RotatE).
More importantly, as an embedding-based framework of Temporal Knowledge Graph Extrapolation, MTKGE could continuously take advantage of the development in the area of temporal knowledge graph embedding, while other state-of-the-art baselines can not.\looseness=-1 
\par
Overall, Table~\ref{table1} and Figure~\ref{comparision} show that our model MTKGE obtains the ability to extrapolate knowledge for unseen entities and relations and outperforms all the baselines through comparison with MaKEr and specifically adapted KGE and TKGE baselines.\looseness=-1 

\subsubsection{\textbf{Findings}}
Comparing all the results of MaKEr and MTKGE in Table~\ref{table1} and Figure~\ref{comparision}, we have two findings.\\

$\bullet$ Finding 1: From Figure~\ref{comparision}, we find that all the best results of MTKGE shown in Table~\ref{table1} come to MTKGE(RotatE), and all the best results of MaKEr come to MaKEr(RotatE). 
RotatE is a KGE model considering relation patterns, including symmetric, anti-symmetric, inverse and composition. This credits to the rich relative position patterns captured by RPPG since the meta-position relations are well encoded and transferred into both seen and unseen relations. RotatE makes full use of them while decoding.\looseness=-1
\par
$\bullet$ Finding 2: We also find that most MTKGE models perform worse on $\text {\emph{u}\_\emph{rel}}$ than on $\text {\emph{u}\_\emph{ent}}$ and $\text {\emph{u}\_\emph{both}}$, which indicates it is still a challenge to extrapolate unseen relation features. 
In IC0515-Ext dataset, the MRR results for MTKGE(RotatE) on $\text {\emph{u}\_\emph{ent}}$ and $\text {\emph{u}\_\emph{both}}$ are around 0.4, but that on $\text {\emph{u}\_\emph{rel}}$ is only 0.3. 
In IC14-Ext dataset, the MRR results on $\text {\emph{u}\_\emph{rel}}$ are still a little lower. 
Despite the difficulties, our proposed MTKGE increases by 96.3\% and 70.7\% on MRR and Hits@10 in IC14-Ext compared with MaKEr in Table~\ref{table1}. \looseness=-1

\subsubsection{\textbf{The Importance of Time in the TSPG Module}}

Compared with models in the extrapolation setting for static KG (i.e., MaKEr, SE-GNN), MTKGE (RotatE) averagely increases by 21\% and 8\% on MRR and Hits@10 for $\text {\emph{u}\_\emph{ent}}$ test quadruples in the two datasets. This confirms that taking time information into account is significant. Furthermore, We raise a question: \emph{Which kind of time information is more beneficial in MTKGE, temporal sequence patterns captured by the TSPG module or explicit timestamp embeddings?}
\par
On the one hand, compared with MTKGE using TKGE as score functions, which more rely on decoding timestamp embeddings, MTKGE using KGE has better performance on extrapolating.
We attribute this phenomenon to the fact that MTKGE has the ability to get temporal meta-knowledge depending on the TSPG module, not the independent timestamps. The meta-time relations gained by the TSPG module are further encoded into relation embeddings. 
On the other hand, we note that the outcomes of Asmp-KGE are also better than that of Asmp-TKGE in the extrapolation setting. We may come to the conclusion that \emph{it is temporal sequence patterns captured by TSPG that have a more vital place}. T-GAP~\citep{TGAP} also has a similar discovery, but they concentrate on concrete temporal displacement.\looseness=-1

\subsection{Ablation Study}

\par
To further emphasize the significance of each module of MTKGE, We conduct an ablation study as follows: 1) only remove the RPPG module; 2) only remove the TSPG module; 3) remove both RPPG module and TSPG module, and replace them with an embedding generation part for unseen relations; 4) replace the entity feature representation module with an embedding generation part for unseen entities; 5) remove GCN for Temporal Knowledge Graph Extrapolation and use embeddings produced by RPPG and TSPG.

\par
The results of ablation studies using MTKGE on IC0515-Ext are shown in Table~\ref{table3}. The results show that removing any part of MTKGE will reduce performance since each module is specifically designed  to produce the embeddings of unseen components. We find that the performance drops the most significantly after removing the two meta-learning modules, limiting the ability of our model to represent unseen components. GCN module also plays a crucial role in extrapolating embeddings for unseen components, and removing it undermines the message passing from the neighbourhood.\looseness=-1

\begin{table}[t]
\centering
\resizebox{\linewidth}{!}{
\begin{tabular}{ccccccc}
\specialrule{0.1em}{1pt}{1pt}
& MTKGE & w/o RPPG & w/o TSPG & w/o Both & w/o EntFeat & w/o GCN\\
\specialrule{0.05em}{1pt}{1pt}
MRR& 0.3564 & 0.2962 & 0.2688 & 0.2384 &0.3205  & 0.2236\\
Hits@1& 25.20 & 20.35  & 15.29  & 12.09  & 23.88  & 10.35\\
\specialrule{0.1em}{1pt}{1pt}
\end{tabular}}
\caption{\label{table3}Ablation study of MTKGE on IC0515-Ext. \emph{w/o Both} is for removing RPPG and TSPG together. (\% for Hits@1)}
\vspace{-0.8cm}
\end{table}

\begin{figure}[t]
	\centering
        \vspace{-0.1cm}
	\begin{minipage}{0.49\linewidth}
		\centering
            
		\includegraphics[scale=0.12]{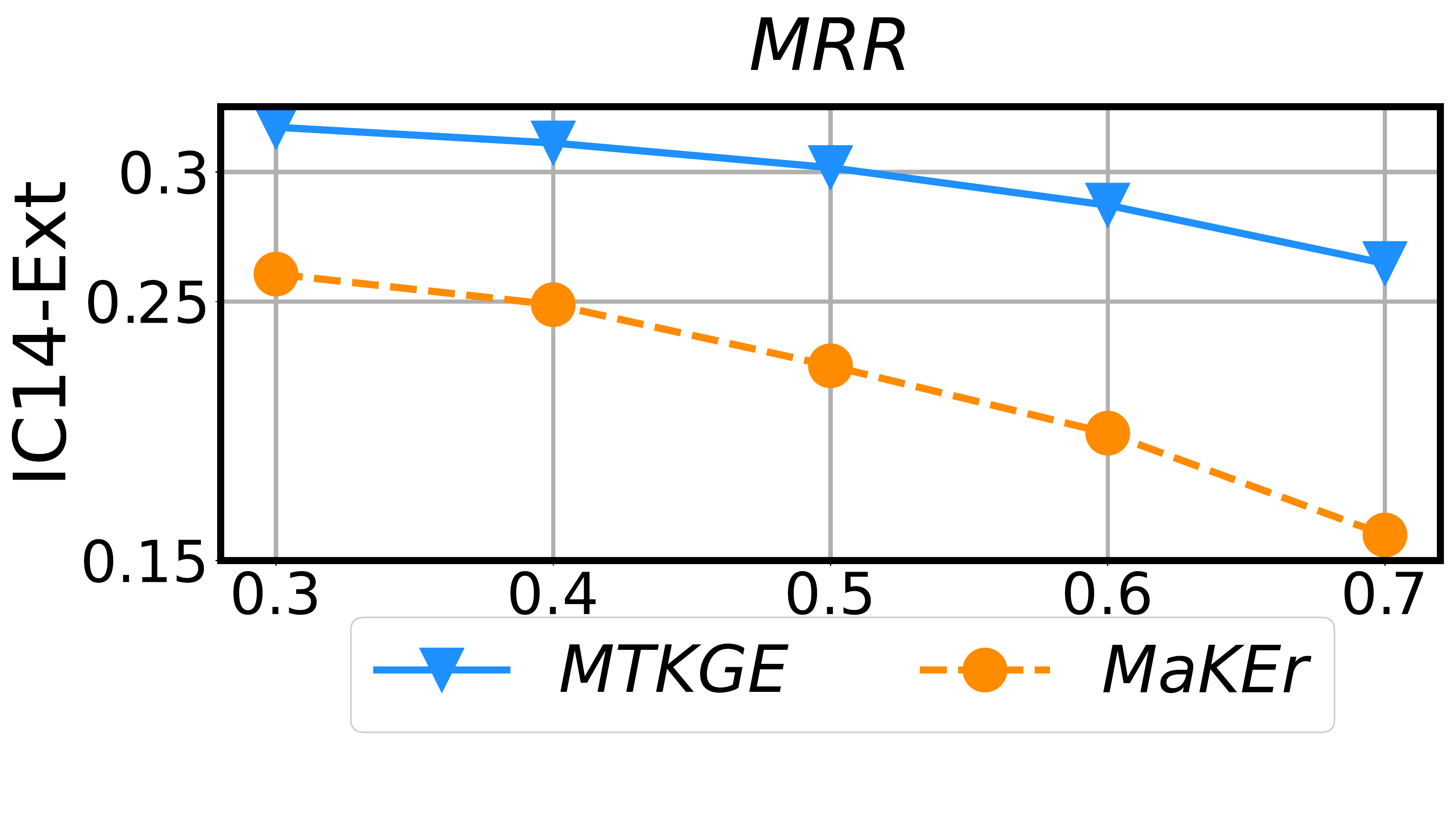}
            \vspace{-0.7cm}
            \hspace{0.3cm plus19mm} 
		\label{chutian1}
	\end{minipage}
	\begin{minipage}{0.49\linewidth}
		\centering
            \vspace{-0.4cm}
             \includegraphics[scale=0.12]{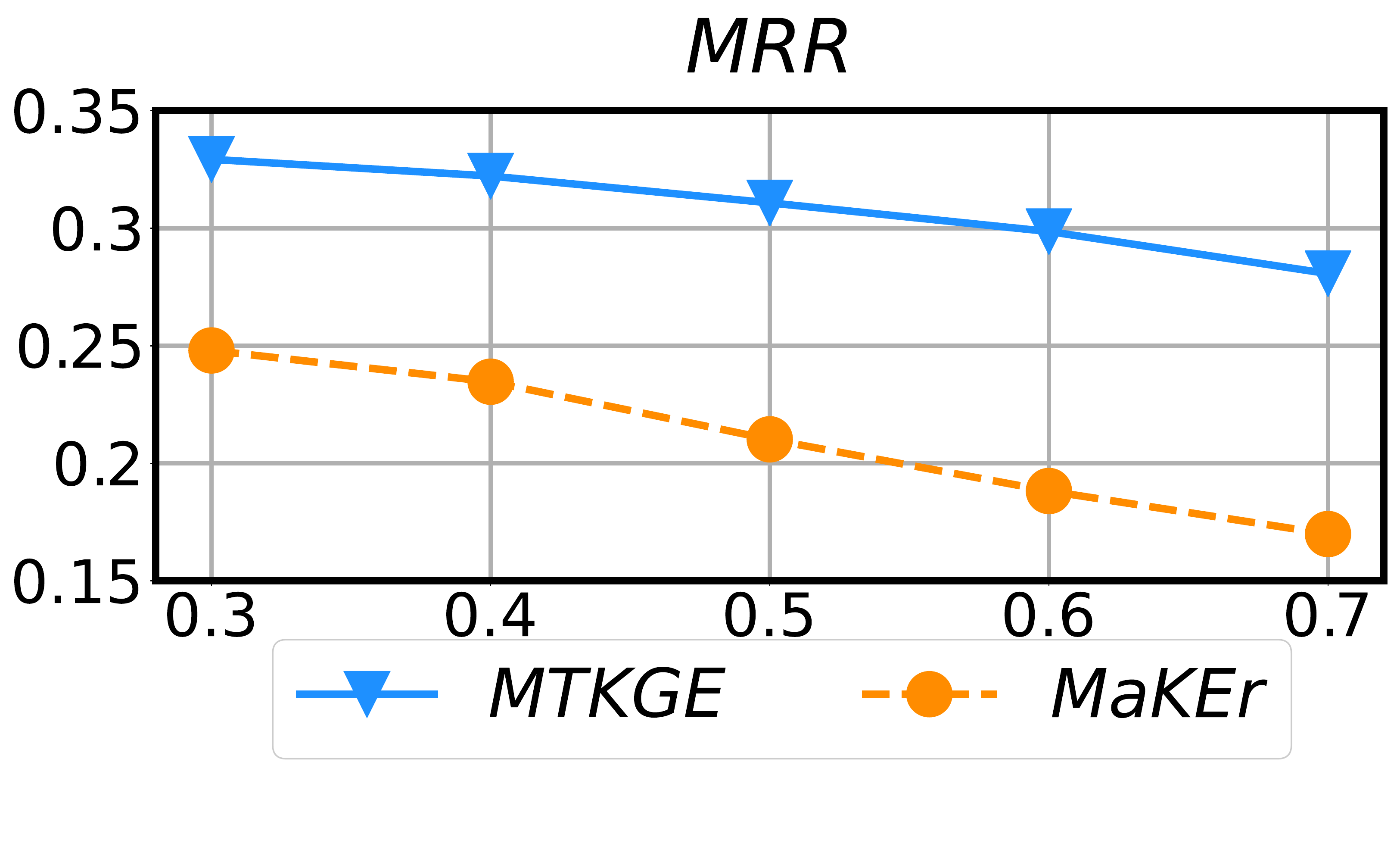}
            \vspace{-0.7cm}
		\label{chutian2}
	\end{minipage}
 \vspace{0.3cm}
        \hspace{-0.03cm }
	\begin{minipage}{0.49\linewidth}
		\centering

            \includegraphics[scale=0.12]{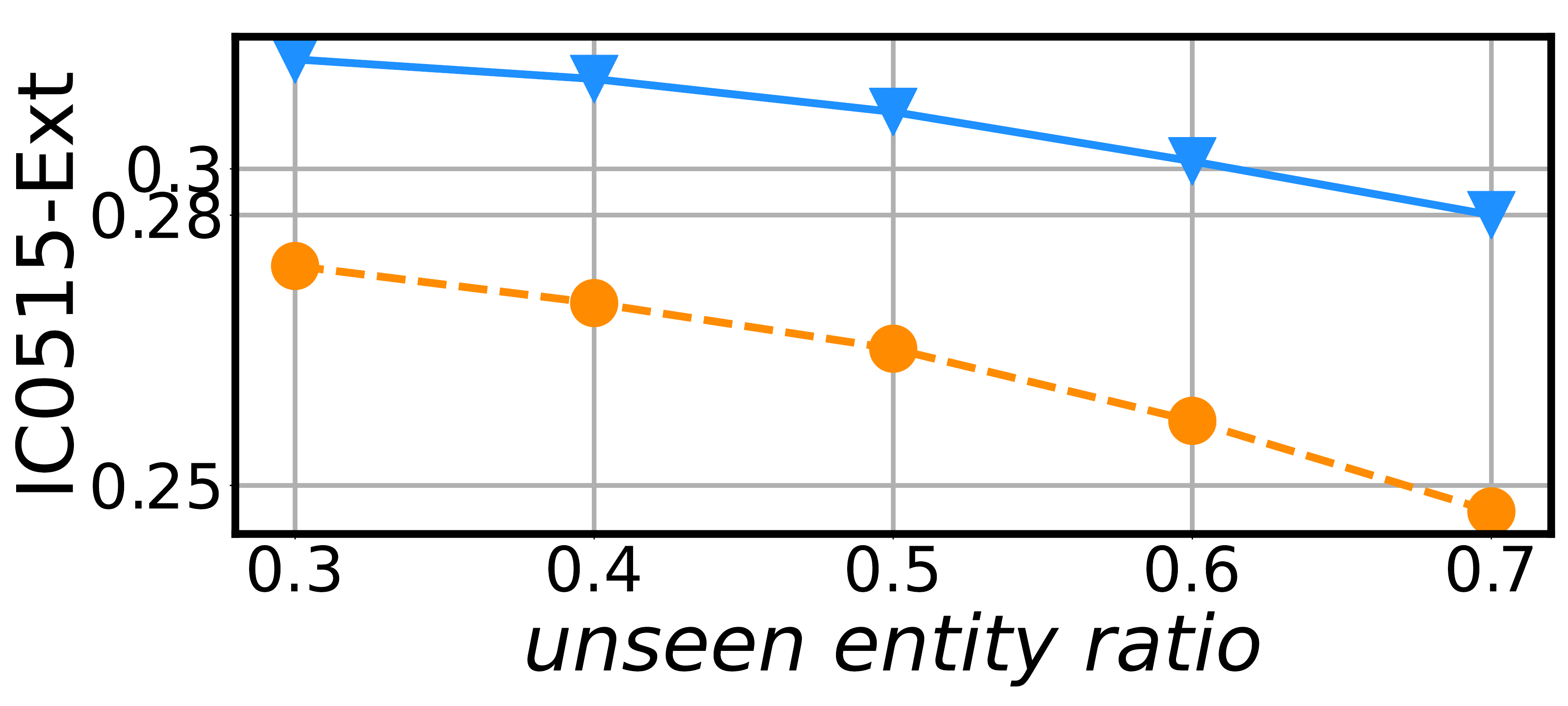}
            \vspace{0cm}
            \hspace{0.8cm} 
		\label{chutian3}
	\end{minipage}
	\begin{minipage}{0.49\linewidth}
		\centering
            \vspace{-0.4cm}
            \hspace{-0.05cm} 	       \includegraphics[scale=0.12]{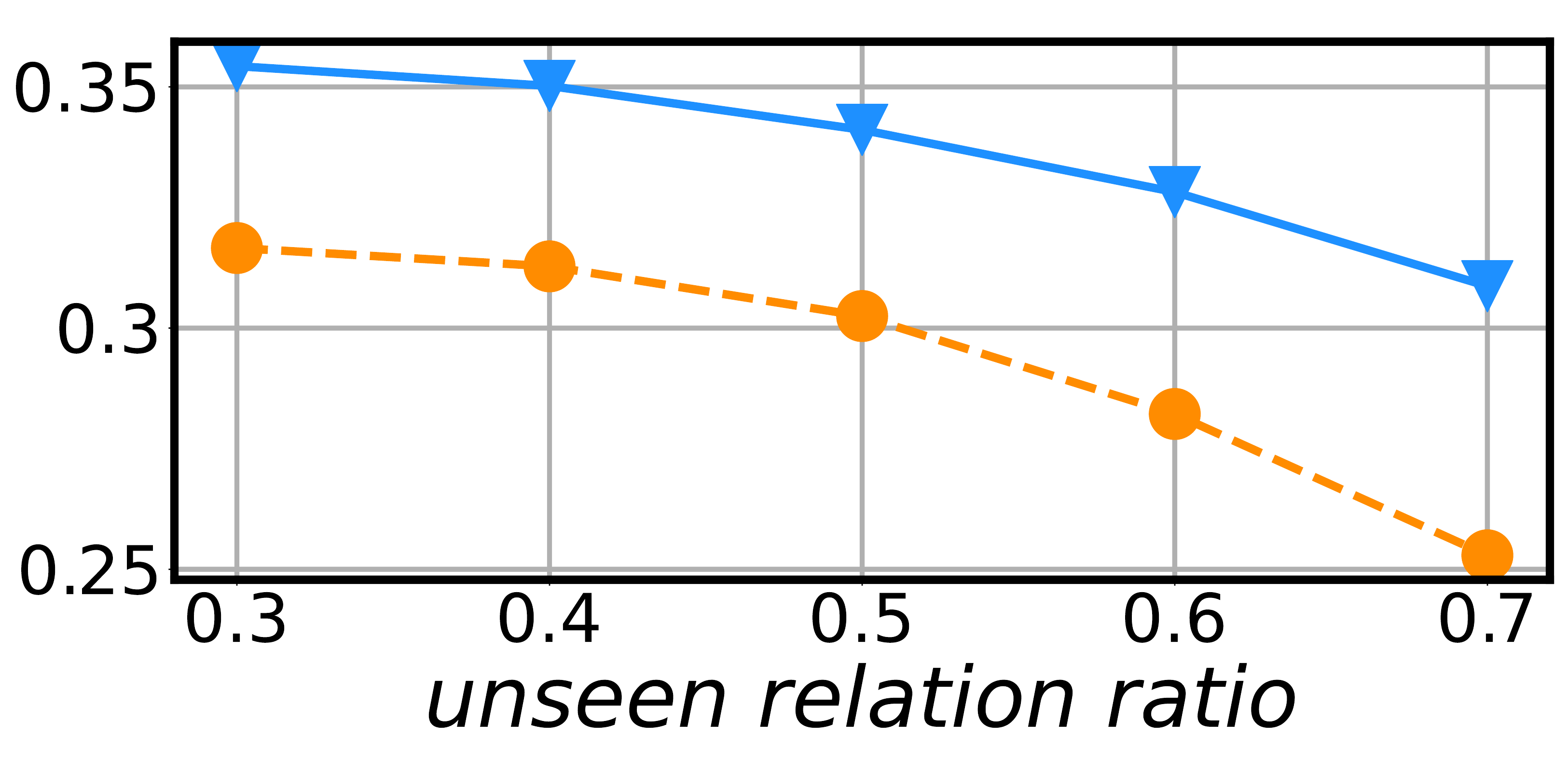}
		\label{chutian4}
	\end{minipage}
 \vspace{-0.7cm}
     \caption{\label{line}MRR results for MTKGE and MaKEr on two datasets simulating various unseen entity ratios (two on the left) and unseen relation ratios (two on the right). The unseen ratios range from 0.3 to 0.7.}
     \vspace{-0.4cm}
\end{figure}

\begin{table*}[t]
\centering

\begin{tabular}{ccc}
\specialrule{0.1em}{1pt}{1pt}
Seen relations  & Unseen relations  & Similarities(MTKGE)\\
\specialrule{0.05em}{1pt}{1pt}
\textcolor{white}{(123)}Cooperate economically \textcolor{white}{(123)}
&\multicolumn{1}{c}{\textcolor{white}{(123)}Provide economic aid \textcolor{white}{(123)}
}&\multicolumn{1}{c}{0.86} \\
&\multicolumn{1}{c}{\textcolor{white}{(123)}Express intent to cooperate\textcolor{white}{(123)}
}&\multicolumn{1}{c}{0.33}\\
\specialrule{0.05em}{1pt}{1pt}
\textcolor{white}{(123)}Make a visit\textcolor{white}{(123)}
&\multicolumn{1}{c}{\textcolor{white}{(123)}Host a visit\textcolor{white}{(123)}
}&\multicolumn{1}{c}{0.81}\\
&\multicolumn{1}{c}{\textcolor{white}{(123)}Call for support\textcolor{white}{(123)}
}&\multicolumn{1}{c}{0.66}\\
&\multicolumn{1}{c}{\textcolor{white}{(123)}Use conventional military force\textcolor{white}{(123)}
}&\multicolumn{1}{c}{-0.04}\\
&\multicolumn{1}{c}{\textcolor{white}{(123)}Meet at a `third' location\textcolor{white}{(123)}
}&\multicolumn{1}{c}{-0.33}\\

\specialrule{0.1em}{1pt}{1pt}
\end{tabular}
\caption{\label{casestudy}Cosine similarities of MTKGE between some seen and unseen relations in several sampled tasks of IC14-Ext dataset.}
\end{table*}

\begin{table*}[t]
\centering
\vspace{-0.3cm}
\begin{tabular}{cccc}
\specialrule{0.1em}{1pt}{1pt}
&temporal query: (Japan, $?$, China, 2014/11/25 )&&\\
&{\color{red}answer: Provide economic aid}&&\\
\specialrule{0.05em}{1pt}{1pt}
\specialrule{0.05em}{1pt}{1pt}
model & quadruples & score & rank\\
\specialrule{0.05em}{1pt}{1pt}
\multirow{2}{*}{\textcolor{white}{(123)}MTKGE\textcolor{white}{(123)}}&\multicolumn{1}{c}{\textcolor{white}{(123)}(Japan, {\color{red}Provide economic aid}, China, 2014/11/25 ) \textcolor{white}{(123)}}&\multicolumn{1}{c}{0.92} &\multicolumn{1}{c}{{\color{red}1}} \\

&\multicolumn{1}{c}{\textcolor{white}{(123)}(Japan, Express intent to cooperate, China, 2014/11/25 ) \textcolor{white}{(123)}}&\multicolumn{1}{c}{0.40}&\multicolumn{1}{c}{14}\\

\specialrule{0.05em}{1pt}{1pt}
\multirow{2}{*}{\textcolor{white}{(123)}MaKEr\textcolor{white}{(123)}}&\multicolumn{1}{c}{\textcolor{white}{(123)}(Japan, {\color{red}Provide economic aid}, China, 2014/11/25 )\textcolor{white}{(123)}}&\multicolumn{1}{c}{0.31}&\multicolumn{1}{c}{{\color{red}12}}\\
&\multicolumn{1}{c}{\textcolor{white}{(123)}(Japan, Express intent to cooperate, China, 2014/11/25 )\textcolor{white}{(123)}}&\multicolumn{1}{c}{0.82}&\multicolumn{1}{c}{1}\\
\specialrule{0.1em}{1pt}{1pt}
\end{tabular}
\caption{\label{example}A temporal link prediction example. Our model MTKGE ranks the ground truth 1st, but MaKEr ranks it 12th.}
\vspace{-0.3cm}
\end{table*}

\subsection{Robustness Study} 
\label{Robustness Study}
\par
IC14-Ext and IC0515-Ext are time-evolving datasets where new entities and relations are added along with emerging facts. Thus, the proportions of new entities and relations in different time periods are different.
Although we cannot control the unseen ratios in the real world, we can split the dataset manually by treating a certain proportion of components as unseen to imitate the actual cases.
\par
Figure~\ref{line} reports the average performance  of  MTKGE and baseline MaKEr  under unseen entity and unseen relation ratios varying from 0.3 to 0.7 with a step size of 0.1 on IC14-Ext and IC0515-Ext. 
First, we can find that all the MRR results of MTKGE are better than 0.25 (0.28) in IC14-Ext (IC0515-Ext), which means the performance of MTKGE is acceptable even at a large proportion of emerging components. 
Secondly, MTKGE is not only superior to MaKEr under all unseen component ratios, but also has a more gradual slope curve. These results demonstrate that our model has a stronger ability to discover meta-knowledge between relations due to  relative position patterns in RPPG and temporal sequence patterns in TSPG, and it is promising for our model to have
an excellent capability of generalization under various real-world situations.\looseness=-1 

\subsection{Degradation From Extrapolation}
In this section, we explore a question:" how much degradation is observed due to the extrapolation?".
As we mentioned above, KGE and TKGE baselines are not designed for Temporal Knowledge Graph Extrapolation. 
Take IC0515-Ext for example, the MRR results of fully trained TComplex, TDistMult and TeRo with no unseen entities are 0.55, 0.45, 0.54. However, the MRR results of their specifically adapted versions are 0.1065, 0.1923 and 0.1555 as shown in Table~\ref{table1}. There is a huge decrease because of the untrained emerging entities and relations.
In contrast, the MRR results of our proposed MTKGE with 0\%, 10\%, 20\% and 30\% of unseen entities are 0.56, 0.53, 0.49 and 0.43. We can see that as the proportion of emerging entities and relationships continues to increase, they slightly reduce the performance of MTKGE. But the results of MTKGE still maintain relatively high until the unseen entity ratio 20\%.
Therefore, the degradation of existing KGE and TKGE models is beyond tolerated and the results of MTKGE do not degrade significantly, which shows the outperforming ability of MTKGE for knowledge extrapolation for temporal knowledge graphs.

\subsection{Case Study} 
\par
IC14-Ext and IC0515-Ext are both datasets with political-type events, where all facts are clearly time-stamped in days. There are interlocking and complex developing patterns among multiple events, such as causality relationships. Individual relations should be seen in the time context, named temporal sequence patterns in our paper, to reflect their semantic meanings. To verify whether MTKGE understands meta-time relations, we calculate the similarities between seen and unseen relations in Table~\ref{casestudy}. 

The higher cosine similarities of two embeddings reflect more vital temporal sequence patterns.\looseness=-1 
\par
In Table~\ref{casestudy}, 
\emph{Cooperate economically} often happens a few days after
\emph{Express intent to cooperate} and then leads to \emph{Provide economic aid} in the future, having a solid temporal sequence. The TSPG module of MTKGE can capture this and show a high positive score, but MaKEr only learns the relative position. 

We show a temporal link prediction example (Japan, $?$, China, 2014/11/25) with ground truth \emph{Provide economic aid} in Table~\ref{example}, MTKGE gives a score of 0.92 to the correct answer (Japan, \emph{Provide economic aid}, China, 2014/11/25) ranking 1st and 0.40 to the confusing candidate. In contrast, MaKEr ranks the correct answer 12th because of not capturing the happening order of these relations.
\emph{Make a visit} is the inverse action of \emph{Host a visit}, so they mostly occur meanwhile. Thus, their similarities are very high in MTKGE, since the two relations are similar in terms of both time and space. Temporally, the aim of a visit may be asking for others' support, so the relation \emph{Call for support} usually takes place after \emph{Make a visit}; this temporal relationship of the two relations also leads to a high score in MTKGE. The temporal logical clues in these cases are leveraged to produce embeddings for unseen entities and relations, achieving extrapolating in TKGs.\looseness=-1

\section{Conclusion}
\par
In this paper, we pose the problem of temporal link prediction with unseen entities and relations in emerging TKGs in the extrapolation setting, called Temporal Knowledge Graph Extrapolation, and propose MTKGE to address this challenge.
Based on the meta-learning approach and GNN framework, MTKGE meta-trains the features of relative position patterns and temporal sequence patterns between relations on a set of sampled tasks and transfers the learned meta-knowledge to unseen components, enhancing their semantic characters. 
Results of extensive experiments show the robust  ability of MTKGE to extrapolate emerging components and the outperforming performance of MTKGE over all the baselines. 
Our future work might further introduce a rule-based approach to construct more interpretable patterns between relations.
\looseness=-1

\bibliographystyle{ACM-Reference-Format}
\bibliography{mybib}

\newpage
\appendix
\section{Model Learning}
\label{loss}
We use self-adversarial negative sampling loss Eq.~\ref{rotateloss} proposed by~\citep{RotatE} to optimize our model, where $p\left(s_{i}^{\prime}, r, o_{i}^{\prime}\right)$ is the self-adversarial weight for a negative
triple among a set of negative samples, and it is calculated by:
\begin{equation}
\begin{aligned}
\begin{split}
p\left(s_{j}^{\prime}, r, o_{j}^{\prime},t\right)=\frac{\exp \alpha \varphi\left(s_{j}^{\prime}, r, o_{j}^{\prime},t\right)}{\sum_{i} \exp \alpha \varphi\left(s_{i}^{\prime}, r, o_{i}^{\prime},t\right)}
\end{split}
\end{aligned}
\end{equation}
where $\alpha$ is the temperature of sampling.

\section{Generate Datasets by Task Sampling}
\label{A}
First, we randomly choose a set of entities, from which we conduct $l_{1}$-length random walk without time constraint to get an expanded entity set. Secondly, we extract all quadruples among these entities in $ {G}$ to form the quadruples for the test TKG, and then remove such quadruples. To make sure that the test TKG exists unseen entities and relations in training KG, we also remove a part of entities and relations from the origin TKG with a random ratio. The validation TKG is extracted in the same way as the test TKG. Thirdly, we sample the training TKG on the remaining TKG $ {G}'$  by choosing a set of entities from $ {G}'$, and conducting $l_{2}$-length random walk from these entities to get an expanded entity set. All the quadruples between these entities are extracted for the training TKG. Finally, we re-label the entities in test KGs in an arbitrary order, as they are viewed as unseen entities.\looseness=-1

\section{Details of A\lowercase{smp}-KGE and A\lowercase{smp}-TKGE}
\label{E}
The calculations for Asmp-TKGE are based on TKGE methods, and there are three embedding calculation operations for each TKGE method: 1) $f_{\textbf{srt} 2 \textbf{o}}$, calculating the tail embedding based on the head, relation and timestamp embeddings; 2) $f_{\textbf{ort} 2 \textbf{s}}$
calculating the head embedding based on the tail, relation and timestamp embeddings; 3) $f_{\textbf{sot} 2 \textbf{r}}$ calculating the relation embedding based on the head and tail embeddings.
\par
For an unseen entity \emph{e} in the test TKG $ {G}^{t e}$, based on the support quadruples, we first find all quadruples linked to \emph{e} that other components are seen during training, then we use $f_{\textbf{ort} 2 \textbf{s}}$ or $f_{\textbf{srt} 2 \textbf{o}}$ to calculate its embedding and take an average. For an unseen relation, we use $f_{\textbf{sot} 2 \textbf{r}}$ to get the embedding in the similar way. For unseen entities and relations which have no linking, we use the average embeddings for all entities and relations as their embeddings. We show the detailed calculations for Asmp-TKGE as follows. \textbf{s}, \textbf{r}, \textbf{o}, \textbf{t} denotes the embeddings for a quadruple (\emph{s}, \emph{r}, \emph{o}, \emph{t}).
\par
TComplEx:
\begin{equation}
\begin{aligned}
f_{\textbf{srt} 2 \textbf{o}}(\textbf{s}, \textbf{r}, \textbf{t}) &=\textbf{s} \circ \textbf{r} \circ \textbf{t}, \\
f_{\textbf{ort} 2 \textbf{s}}(\textbf{o}, \textbf{r}, \textbf{t}) &=\overline{\overline{\textbf{o}} \circ \textbf{r} \circ \textbf{t}}, \\
f_{\textbf{sot} 2 \textbf{r}}(\textbf{s}, \textbf{o}, \textbf{t}) &=\overline{\textbf{s} \circ \overline{\textbf{o}} \circ \textbf{t}}.
\notag
\end{aligned}
\end{equation}

\par
TDistMult:
\begin{equation}
\begin{aligned}
f_{\textbf{srt} 2 \textbf{o}}(\textbf{s}, \textbf{r}, \textbf{t}) &=\textbf{s} \circ \textbf{r} \circ \textbf{t}, \\
f_{\textbf{ort} 2 \textbf{s}}(\textbf{o}, \textbf{r}, \textbf{t}) &=\textbf{o} \circ \textbf{r} \circ \textbf{t}, \\
f_{\textbf{sot} 2 \textbf{r}}(\textbf{s}, \textbf{o}, \textbf{t}) &=\textbf{s} \circ \textbf{o} \circ \textbf{t}.
\notag
\end{aligned}
\end{equation}

\par
TeRo:
\begin{equation}
\begin{aligned}
f_{\textbf{srt} 2 \textbf{o}}(\textbf{s}, \textbf{r}, \textbf{t}) &=\overline{(\textbf{s} \circ \textbf{t} + \textbf{r})} / \textbf{t}  ,\\
f_{\textbf{ort} 2 \textbf{s}}(\textbf{o}, \textbf{r}, \textbf{t}) &=(\textbf{o} \circ \textbf{t} - \textbf{r} )/ \textbf{t}, \\
f_{\textbf{sot} 2 \textbf{r}}(\textbf{s}, \textbf{o}, \textbf{t}) &=\overline{\textbf{o} \circ \textbf{t}} - \textbf{s} \circ \textbf{t} .
\notag
\end{aligned}
\end{equation}
\par
In Asmp-KGE, \textbf{s}, \textbf{r}, \textbf{o} denotes the embeddings for a quadruples (\emph{s}, \emph{r}, \emph{o}, \emph{t}), we skip the time factor:
\par
ComplEx:
\begin{equation}
\begin{aligned}
f_{\textbf{sr} 2 \textbf{o}}(\textbf{s}, \textbf{r}) &=\textbf{s} \circ \textbf{r}, \\
f_{\textbf{or} 2 \textbf{s}}(\textbf{o}, \textbf{r}) &=\overline{\overline{\textbf{o}} \circ \textbf{r}}, \\
f_{\textbf{so} 2 \textbf{r}}(\textbf{s}, \textbf{o}) &=\overline{\textbf{s} \circ \overline{\textbf{o}}}.
\notag
\end{aligned}
\end{equation}

\par
DistMult:
\begin{equation}
\begin{aligned}
f_{\textbf{sr} 2 \textbf{o}}(\textbf{s}, \textbf{r}) &=\textbf{s} \circ \textbf{r}, \\
f_{\textbf{or} 2 \textbf{s}}(\textbf{o}, \textbf{r}) &=\textbf{o} \circ \textbf{r}, \\
f_{\textbf{so} 2 \textbf{r}}(\textbf{s}, \textbf{o}) &=\textbf{s} \circ \textbf{o}.
\notag
\end{aligned}
\end{equation}

\par
RotatE:
\begin{equation}
\begin{aligned}
f_{\textbf{sr} 2 \textbf{o}}(\textbf{s}, \textbf{r}) &=\textbf{s} \circ \textbf{r}, \\
f_{\textbf{or} 2 \textbf{s}}(\textbf{o}, \textbf{r}) &=\textbf{o} / \textbf{r}, \\
f_{\textbf{so} 2 \textbf{r}}(\textbf{s}, \textbf{o}) &=\textbf{o} / \textbf{s}.
\notag
\end{aligned}
\end{equation}

\section{MTKGE and MaKEr with six score functions}
\label{D}
\begin{figure*}[htbp]
	\centering
	\subfigure[MRR on unseen entities($\text{\emph{u}\_\emph{ent}}$)]{
		\begin{minipage}[t]{0.33\linewidth}
			\centering
			\includegraphics[width=2.3in]{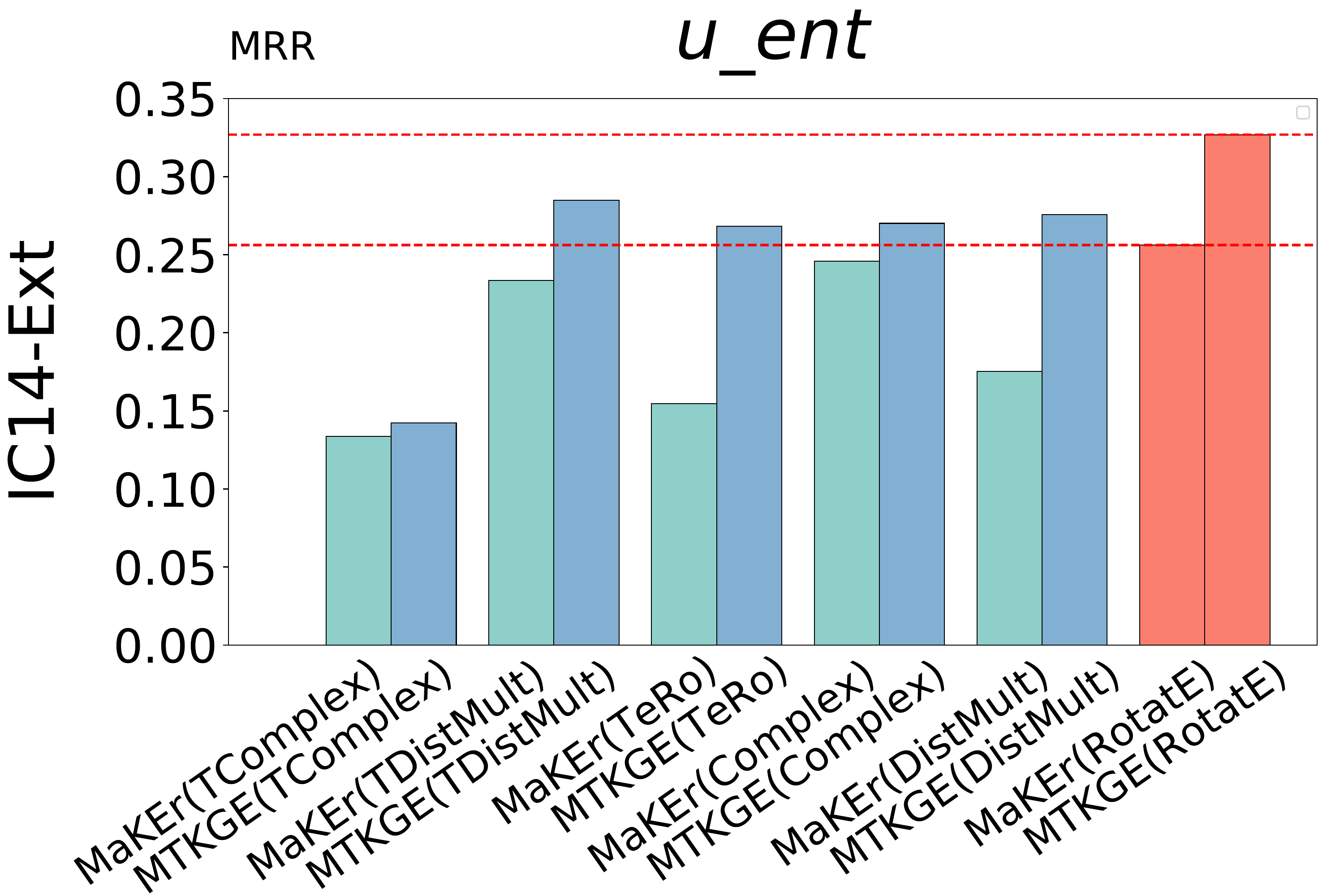}\\
			\vspace{0.15cm}
			\includegraphics[width=2.3in]{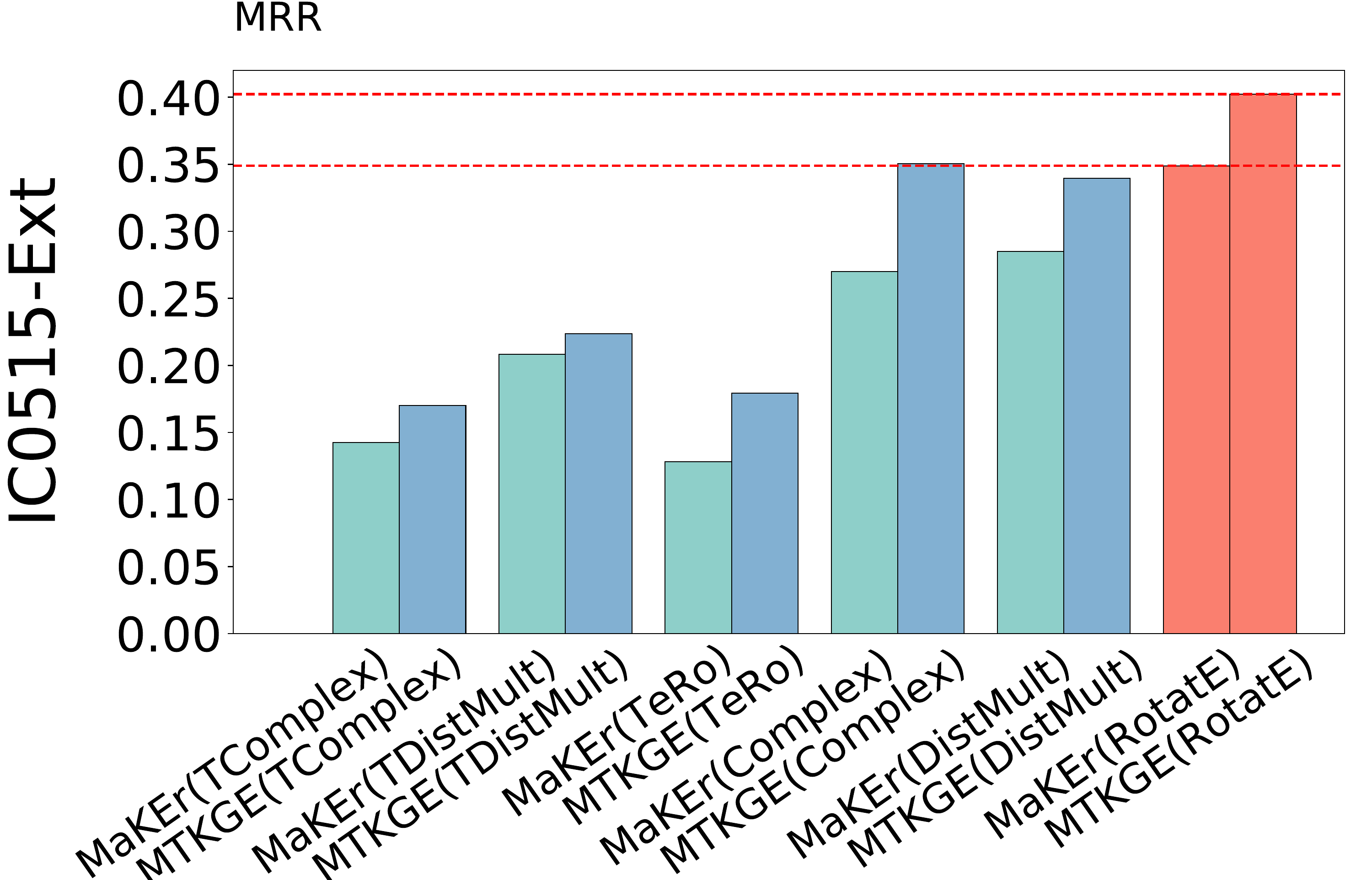}\\
			\vspace{0.06cm}
		\end{minipage}%
	}%
	\subfigure[MRR on unseen relations($\text{\emph{u}\_\emph{rel}}$)]{
		\begin{minipage}[t]{0.33\linewidth}
			\centering
			\includegraphics[width=2.3in]{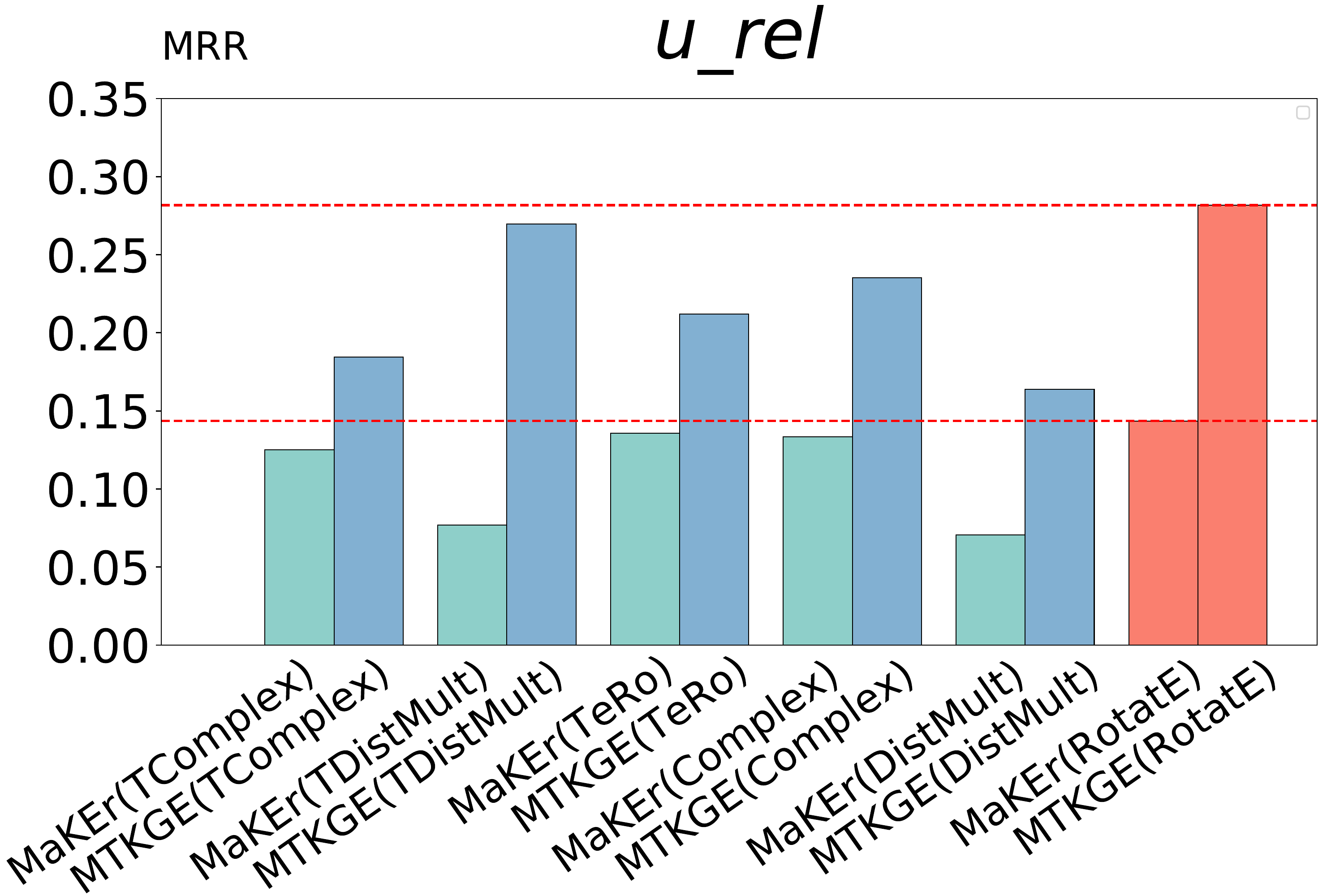}\\
			\vspace{0.02cm}
			\includegraphics[width=2.3in]{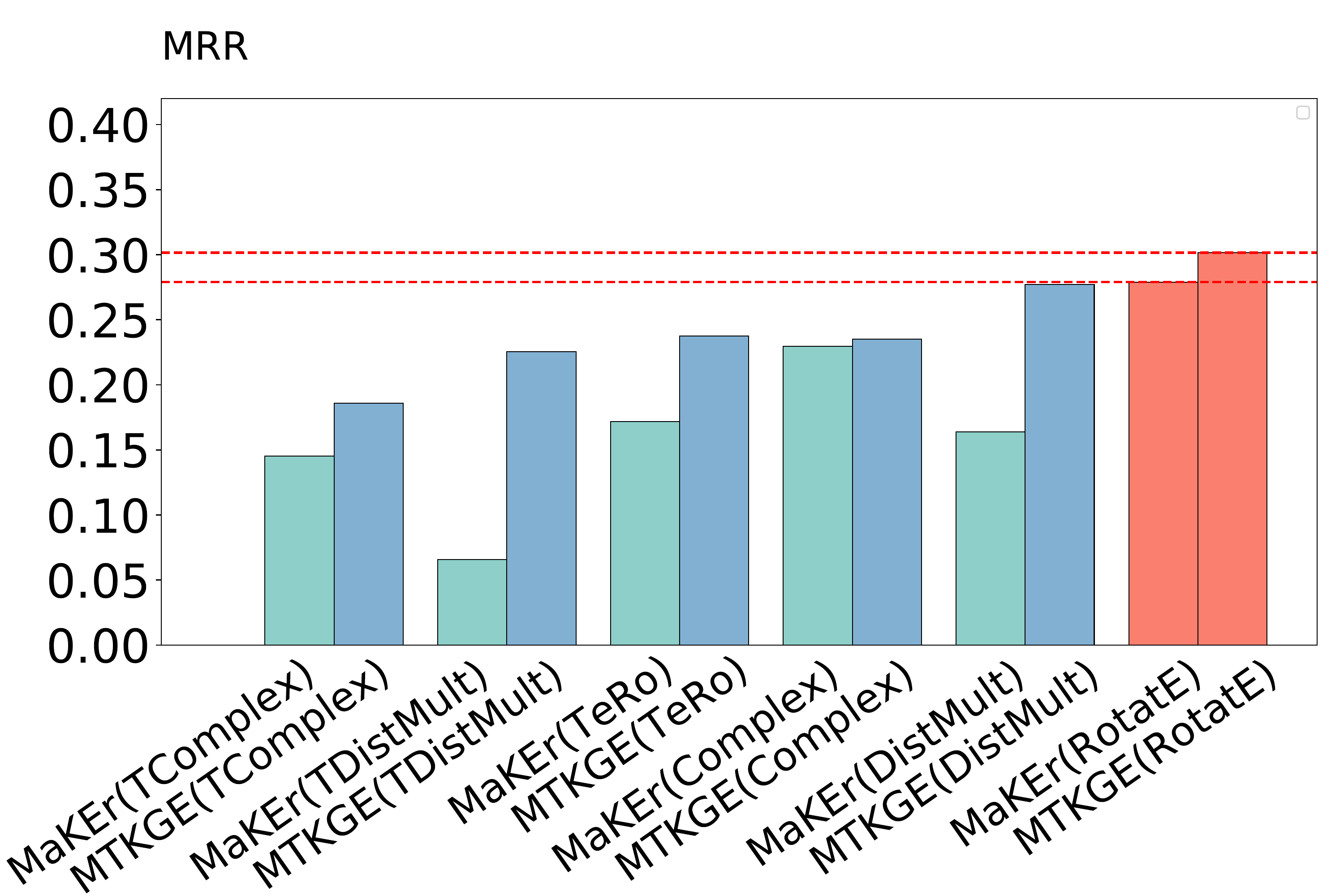}\\
			\vspace{0.02cm}
		\end{minipage}%
	}%
	\subfigure[MRR on unseen entities and relations($\text{\emph{u}\_\emph{both}}$)]{
		\begin{minipage}[t]{0.33\linewidth}
			\centering
			\includegraphics[width=2.3in]{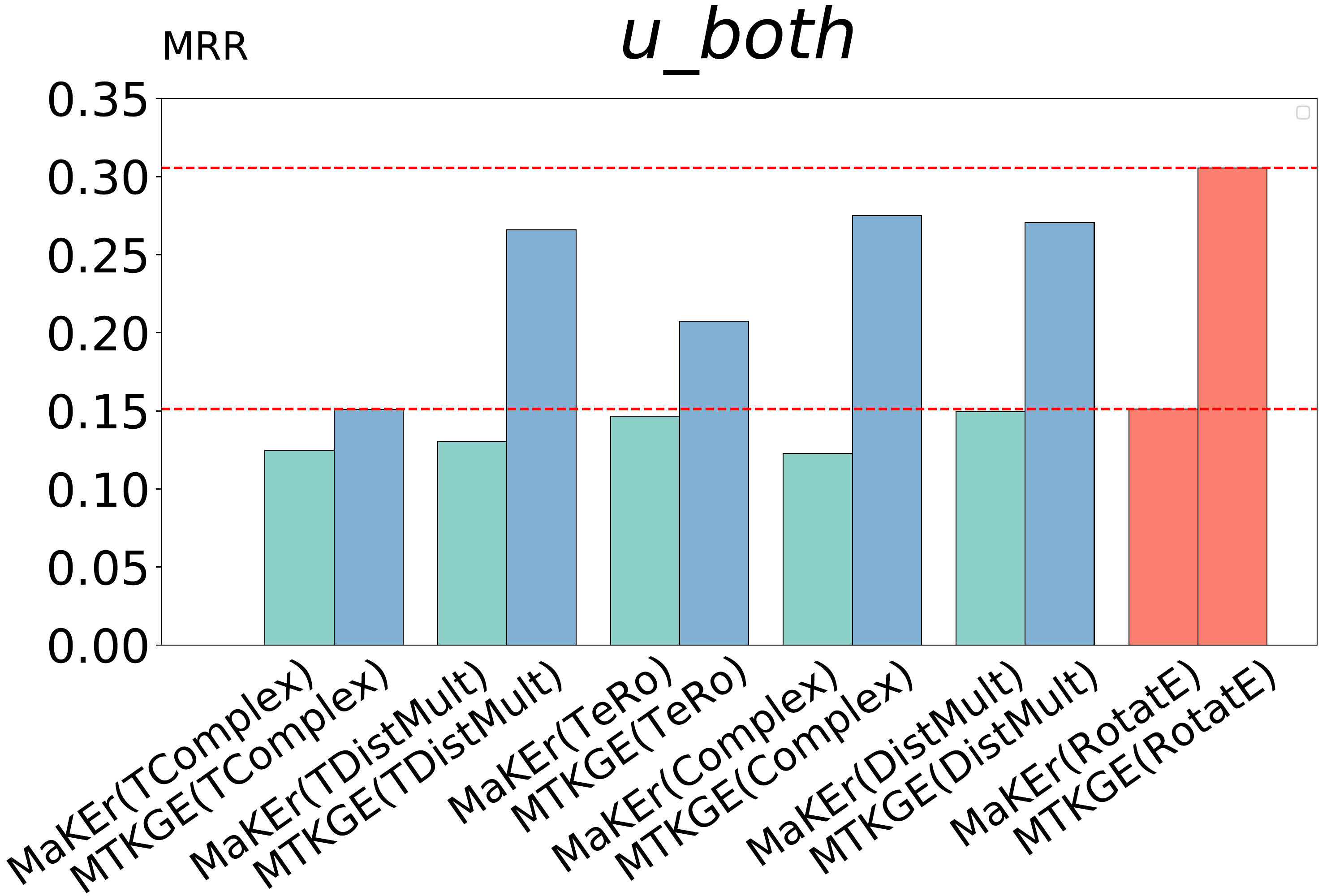}\\
			\vspace{0.02cm}
			\includegraphics[width=2.3in]{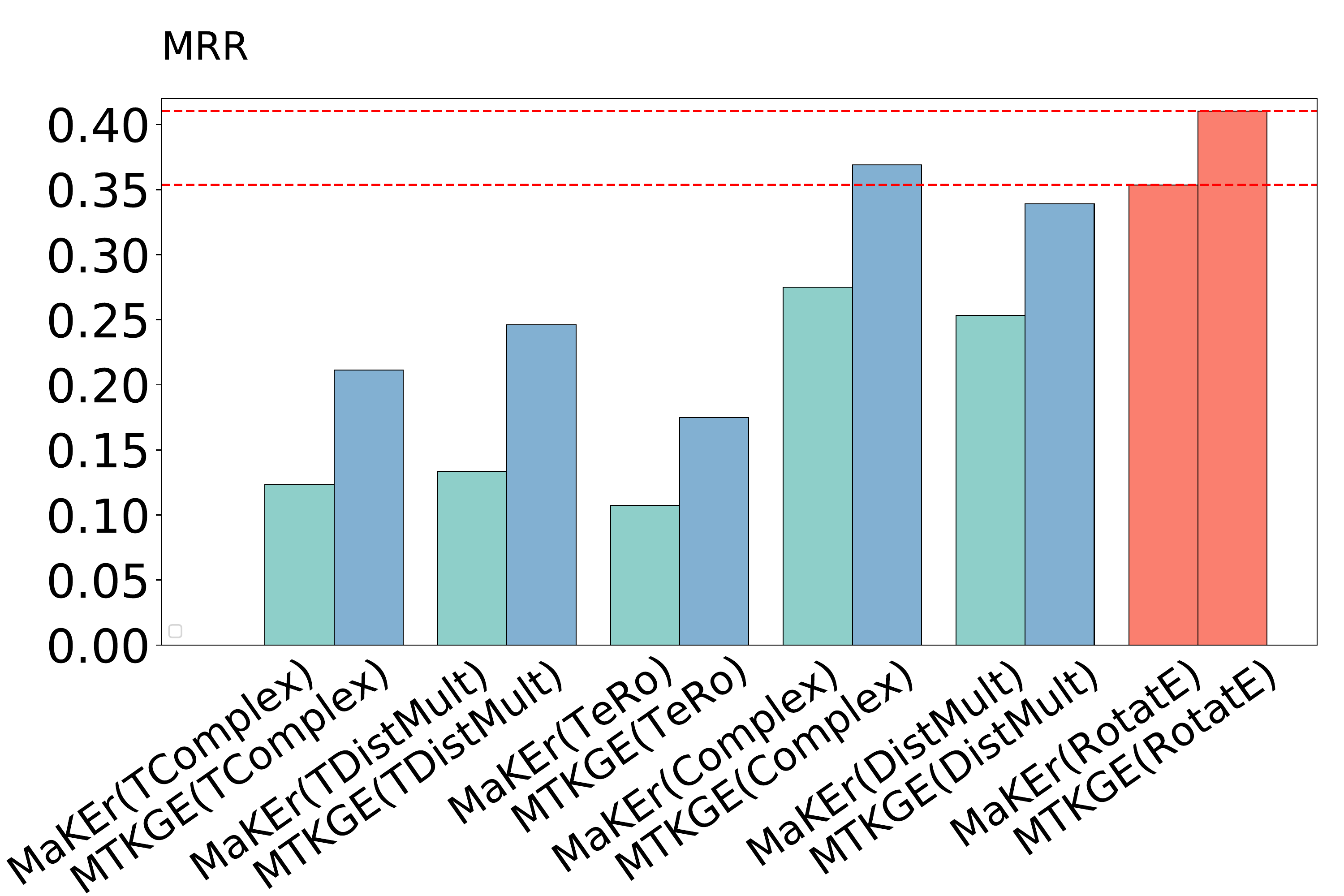}\\
			\vspace{0.02cm}
		\end{minipage}%
	}%
	\centering
	\caption{Link prediction results (MRR) of MaKEr and our model MTKGE using various score functions for three kinds of query quadruples ($\text{\emph{u}\_\emph{ent}}$, $\text{\emph{u}\_\emph{rel}}$, $\text{\emph{u}\_\emph{both}}$) on two datasets. MaKEr(RotatE) means MaKEr using RotatE as the score function, MTKGE(RotatE) means MTKGE using RotatE as the score function, and the same for other score functions.}
	\label{comparision}
\end{figure*}

Figure~\ref{comparision} shows the results of MTKGE and MaKEr with six score functions.

\end{document}